\documentclass[10pt,twocolumn,letterpaper]{article}

\usepackage[pagenumbers]{cvpr} 

\usepackage{graphicx}
\usepackage{amsmath}
\usepackage{amssymb}
\usepackage{booktabs}
\usepackage{xcolor}
\usepackage{bm}
\usepackage[pagebackref,breaklinks,colorlinks]{hyperref}
\usepackage{enumitem}
\usepackage{multirow}

\usepackage{pdfrender}
\newcommand*{\boldcheckmark}{%
	\textpdfrender{
		TextRenderingMode=FillStroke,
		LineWidth=.5pt, 
	}{\checkmark}%
}
\usepackage{colortbl} 
\definecolor{mygray-bg}{gray}{0.9}
\usepackage{arydshln} 

\usepackage[capitalize]{cleveref}
\crefname{section}{Sec.}{Secs.}
\Crefname{section}{Section}{Sections}
\Crefname{table}{Table}{Tables}
\crefname{table}{Tab.}{Tabs.}

\definecolor{gray}{RGB}{120, 120, 120}
\definecolor{darkred}{RGB}{135, 12, 12}
\definecolor{lightblue}{RGB}{0, 128, 192}

\def\cvprPaperID{6703} 
\def\confName{CVPR}
\def\confYear{2022}

\begin{document}

\title{\emph{Classification-Then-Grounding}: \\
Reformulating Video Scene Graphs as Temporal Bipartite Graphs}


\author{
    Kaifeng Gao$^\dagger$, \;
    Long Chen$^\ddagger$\thanks{Long Chen is the corresponding author. This work started when LC at Zhejiang University, and YN at Nanyang Technological University.}, \;
    Yulei Niu$^\ddagger$, \;
    Jian Shao$^\dagger$, \;
    Jun Xiao$^\dagger$ \;
    \\
    $^\dagger$Zhejiang University, \;\; $^\ddagger$Columbia University \\
 {\tt\small\{kite\_phone,jshao,junx\}@zju.edu.cn} ~  {\tt\small\{zjuchenlong,yn.yuleiniu\}@gmail.com}
}

\maketitle

\begin{abstract}
Today's VidSGG models are all proposal-based methods, i.e., they first generate numerous paired subject-object snippets as proposals, and then conduct predicate classification for each proposal. In this paper, we argue that this prevalent proposal-based framework has three inherent drawbacks: 1) The ground-truth predicate labels for proposals are partially correct. 2) They break the high-order relations among different predicate instances of a same subject-object pair. 3) VidSGG performance is upper-bounded by the quality of the proposals. To this end, we propose a new \textbf{classification-then-grounding} framework for VidSGG, which can avoid all the three overlooked drawbacks. Meanwhile, under this framework, we reformulate the video scene graphs as temporal bipartite graphs, where the entities and predicates are two types of nodes with time slots, and the edges denote different semantic roles between these nodes. This formulation takes full advantage of our new framework. Accordingly, we further propose a novel \textbf{BI}partite \textbf{G}raph based SGG model: \textbf{BIG}. It consists of a classification stage and a grounding stage, where the former aims to classify the categories of all the nodes and the edges, and the latter tries to localize the temporal location of each relation instance. Extensive ablations on two VidSGG datasets have attested to the effectiveness of our framework and BIG. Code is available at \url{https://github.com/Dawn-LX/VidSGG-BIG}.
\end{abstract}

\begin{figure}[t]
  \begin{center}
  \includegraphics[width=\linewidth]{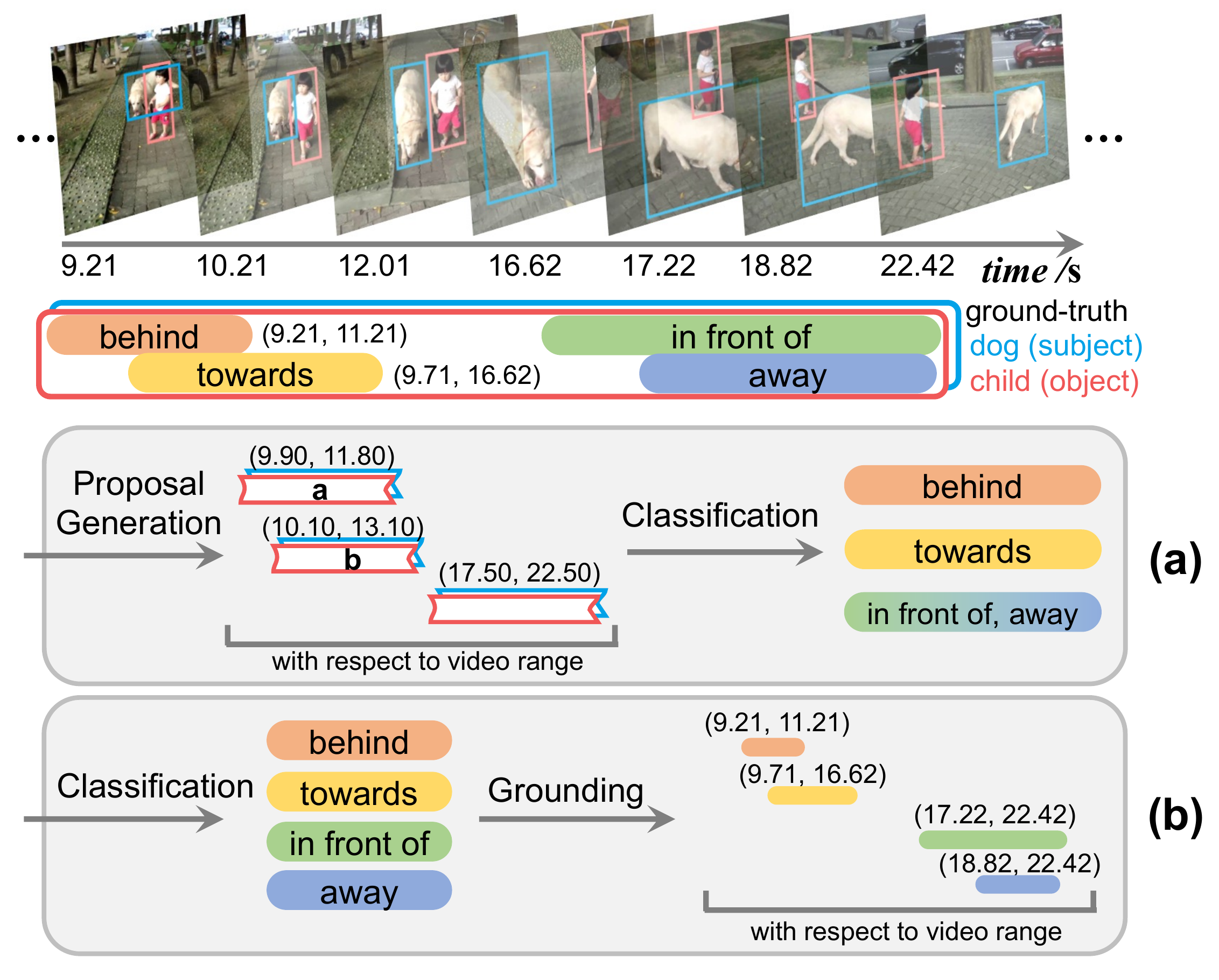}
  \end{center}
  \vspace{-2em}
  \caption{\textbf{(a)}: The pipeline of \textbf{proposal-based} framework. Given a video, it first generates numerous proposals (with different time slots), and then conducts predicate classification for each proposal. \textbf{(b)}: The pipeline of \textbf{classification-then-grounding} framework. It first conducts predicate classification based on the whole tracklet pair, and then grounds all the predicted relation instances.}
  \label{fig:motivation}
\end{figure}

\section{Introduction}
To bridge the gap between vision and other modalities (\eg, language), a surge of interests in our community start to convert the vision data into graph-structured representations, called \textbf{scene graphs}~\cite{johnson2015image}. 
Scene graphs are visually-grounded graphs, where the nodes and edges represent object instances (or entities) and their pairwise visual relations (predicates), respectively. Due to the inherent interpretability, scene graphs have been widely used in numerous downstream tasks to help boost model performance, \eg, captioning~\cite{yang2019auto,chen2021human,chen2017sca}, grounding~\cite{liu2019joint,chen2021ref}, and 
QA~\cite{hudson2019gqa,chen2020counterfactual,chen2021counterfactual,niu2021introspective}.

Video Scene Graph Generation (\textbf{VidSGG)} has achieved significant progress over the recent years. Currently, almost all existing VidSGG models are proposal-based\footnote{We use \emph{proposals} to represent paired subject-object tracklet segments.}. Specifically, they can be categorized into two groups: 1) \emph{Segment-proposal based}: They first cut the video into short segments and detect object tracklets in each segment to compose segment proposals, then classify predicates in each proposal and merge all predicted relation triplets (\ie, $\langle$\texttt{subject}, \texttt{predicate}, \texttt{object}$\rangle$) among adjacent segments~\cite{shang2017video,qian2019video,su2020video}. However, they fail to exploit the long-term context in the video (or tracklets) due to the limits of short segments. 2) \emph{Tracklet-proposal based}: They directly detect tracklets in the whole video and generate tracklet proposals by sliding-windows~\cite{liu2020beyond} or confidence splitting~\cite{Chen2021Social}, and then conduct predicate classification for each proposal.

Although these proposal-based methods have dominated the performance on VidSGG datasets, it is worth noting that this prevalent framework has three inherent drawbacks: 

\textbf{\emph{1. The ground-truth predicate labels for proposals are partially correct}}. By ``partially", we mean that the ground-truth predicate labels sometimes are \emph{WRONG}. Specifically, following the IoU-based strategy in object detection, existing proposal-based models all assign predicate labels to proposals based on the volume IoU (vIoU). This strategy naturally discards some ``ground-truth" predicates if their vIoUs are less than the threshold. As shown in Figure~\ref{fig:motivation}(a), two relations \texttt{behind} and \texttt{towards} happen simultaneously on multiple frames inside both \textsf{proposal}$_a$ and \textsf{proposal}$_b$, but the assigned predicate label for \textsf{proposal}$_a$ is only \texttt{behind} (and \texttt{towards} for \textsf{proposal}$_b$)\footnote{For \textsf{proposal}$_a$, its vIoU with predicate \texttt{towards} $<$ 0.5 and its vIoU with predicate \texttt{behind} $>$ 0.5. The situation is opposite for \textsf{proposal}$_b$.}. Meanwhile, once a predicate label is assigned to the proposal, they assume this relation should last for the whole proposal (\ie, it happens in all the frames of the proposal). Obviously, one of the negative impacts of this issue is that the ground-truth labels for two highly-overlapped proposals (\textsf{proposal}$_{a/b}$) may be totally different, and this inconsistency hurts the model training.

\textbf{\emph{2. They break the high-order relations among different predicate instances of a same subject-object pair}}. Due to the nature of videos, there are always multiple relations happening between a same subject-object pair, and these relations can serve as critical context (or inductive bias) to benefit the predictions of other relations. For example, \texttt{behind}, \texttt{towards}, and \texttt{away} always happen sequentially between 
\texttt{dog} and \texttt{child}. Instead, proposal-based methods explicitly break these high-order relations by pre-cutting tracklets, and classify predicates independently in each proposal\footnote{Although a few proposal-based models start to resort to some context modeling techniques to remedy this weakness, we claim that the proposal-based framework itself overlooks and breaks these high-order relations.}.

\textbf{\emph{3. VidSGG performance is upper-bounded by the quality of the proposals}}. The VidSGG performance is sensitive to the heuristic rules for proposal generation (\eg, the sizes or number of proposals). Meanwhile, to achieve higher recalls, they always generate excessive proposals, which significantly increases the computation complexity.

In this paper, we propose a \emph{classification-then-grounding} framework for VidSGG, which can avoid all the mentioned drawbacks in proposal-based methods. Specifically, we first conduct predicate classification based on the whole tracklets, and then ground each predicted predicate instance (Figure~\ref{fig:motivation}(b)). Compared to proposal-based methods, we regard all the relations happen between the two tracklets as ground-truth predicate labels (\eg, \texttt{behind}, \texttt{towards}, \texttt{away}, and \texttt{in-front-of} are all ground-truth predicates for \texttt{dog} and \texttt{child}). Our framework not only provides more accurate ground-truth predicate labels, but also preserves the ability to utilize high-order relations among predicates. Moreover, it avoids superfluous proposals and heuristic rules.

\begin{figure}[t]
  \centering
  \includegraphics[width=\linewidth]{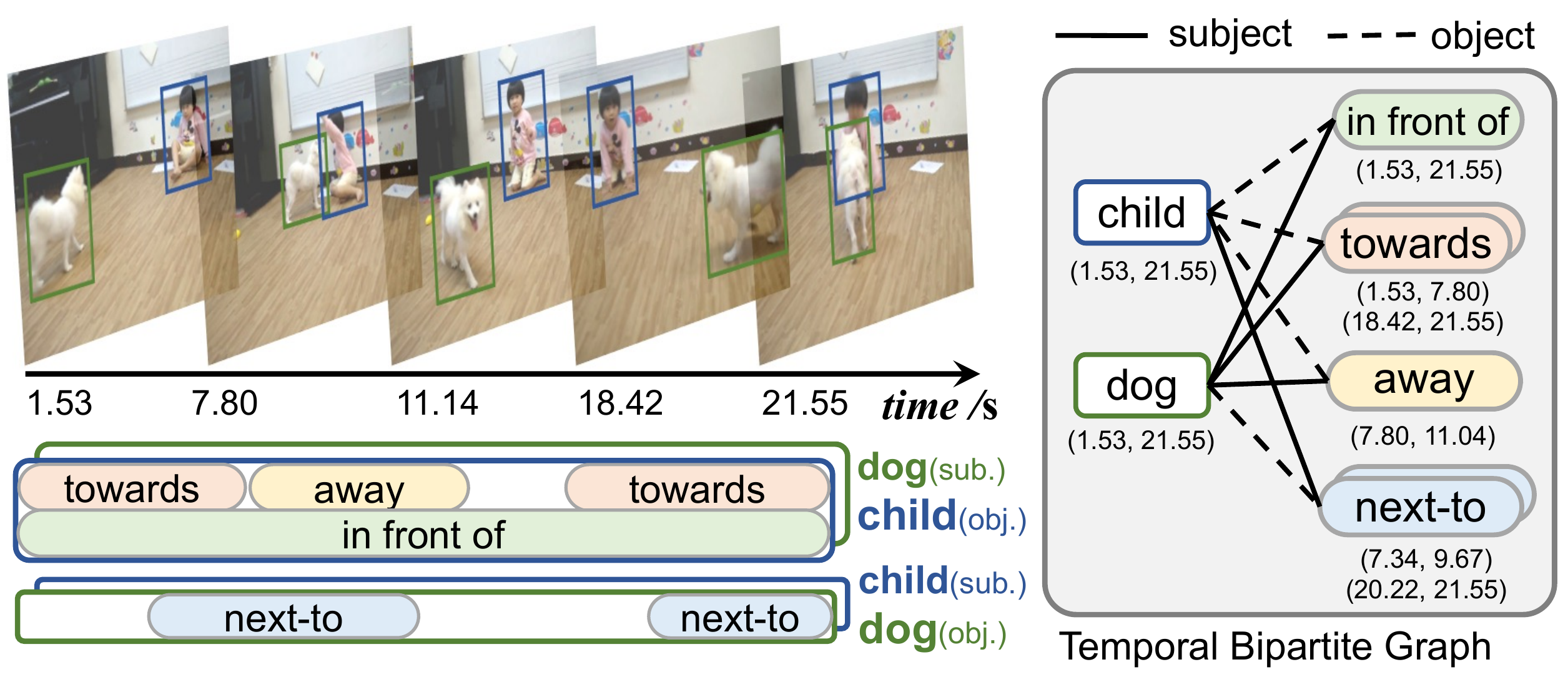}
  \vspace{-2em}
  \caption{\textbf{Left}: A video example and its ground-truth visual relation triplets. \textbf{Right}: The corresponding temporal bipartite graph. Comparisons with existing formulation are left in appendix.}
  \label{fig:bipartite}
\end{figure}

Under this framework, we propose to reformulate video scene graphs as temporal bipartite graphs, where the entities and predicates are two types of nodes with time slots, and the edges denote different semantic roles (\ie, \texttt{subject} and \texttt{object}) between these nodes (Figure~\ref{fig:bipartite}). Each entity node is an object tracklet, and its time slot is the temporal range of this tracklet. Each predicate node is a set of relation \emph{instances} between two entities with the same predicate category, where each time slot denotes the temporal range of each relation instance (\eg, predicate node \texttt{towards} in Figure~\ref{fig:bipartite} has two time slots). Thus, each entity node can be linked with multiple predicate nodes to represent multiple relations involved, and each predicate node can be linked with at most one entity node for each role. This formulation can not only be easily extended to more general relations with more semantic roles~\cite{zareian2020weakly}, but also avoid exhaustively enumerating all entity pairs for predicate prediction.

Accordingly, we propose a \textbf{BI}partite \textbf{G}raph based model \textbf{BIG}, which consists of a \emph{classification} stage and a \emph{grounding} stage. Specifically, the former aims to classify the categories of all the nodes and edges, and the latter tries to localize the temporal location of each relation instance. \textbf{For the classification stage}, it is a Transformer-based model, where the inputs for the encoder and decoder are tracklet features and learnable predicate embeddings, respectively. To distinguish different semantic roles, we also propose a role-aware cross-attention which introduces role-wise distinctions into predicate embeddings. \textbf{For the grounding stage}, we regard the triplet categories of each predicate node as a language query (\eg, $\langle$\texttt{dog}, \texttt{towards}, \texttt{child}$\rangle$ in Figure~\ref{fig:bipartite}), and ground this language query in the video. Since each relation category may happen multiple times between two tracklets, we design a multi-instance grounding head at this stage.

We evaluate models on two challenging VidSGG benchmarks: VidVRD~\cite{shang2017video} and VidOR~\cite{shang2019annotating}. Extensive ablations and results have demonstrated the effectiveness of our new classification-then-grounding framework and BIG model. 

In summary, we make three contributions in this paper:
\vspace{-0.7em}
\begin{enumerate}[leftmargin=4mm]
    \itemsep-0.5em
    \item We propose a new classification-then-grounding framework for VidSGG. It avoids three inherent drawbacks of the existing proposal-based framework.
    
    \item We reformulate video scene graphs as temporal bipartite graphs, and take full advantage of the new framework.
    
    \item We propose a novel model BIG, which achieves state-of-the-art performance on two VidSGG datasets. 
\end{enumerate}

\section{Related Works}

\noindent\textbf{Video Scene Graph Generation.} Today's VidSGG models are all proposal-based. They usually focus on designing: 1) more effective context fusing mechanism among segment or tracklet proposals, \eg, GCNs or CRFs~\cite{liu2020beyond,qian2019video,tsai2019video}, compositional relation encoding~\cite{Chen2021Social}, or structure context aggregation~\cite{teng2021target}; or 2) stronger relation association approaches, such as MHA~\cite{su2020video} or online association~\cite{qian2019video}. In contrast, we are the first to avoid the proposal generation step and solve VidSGG task in a new classification-then-grounding manner. Meanwhile, we propose a temporal bipartite graph formulation by extending the image bipartite graph~\cite{zareian2020weakly} into the video domain, \ie, assigning predicate nodes with time slots. Accordingly, we propose a novel BIG model.

\noindent\textbf{Transformer Structure for SGG.}
Transformer structures~\cite{vaswani2017attention,wang2022crossformer} regain vision community attention after the pioneering work DETR~\cite{carion2020end}, which regards the object detection task as a set prediction problem. Inspired from DETR, several recent works start to use Transformer models for image scene graph generation~\cite{zou2021end,tamura2021qpic,chen2021reformulating,cong2021spatial}. Similarly, these models utilize a set of learnable embeddings as the input of the decoder, and predict the triplets based on the encoded global object features. Inspired from these works, we also adopt the Transformer structure in our classification stage, and design a role-aware cross-attention module to explicitly model different edges of the temporal bipartite graph.

\noindent\textbf{Video Grounding.} 
It aims to localize the video segment depicted by a language query~\cite{gao2017tall,yuan2021closer}. Existing models can be roughly grouped into: 1) Anchor-based~\cite{anne2017localizing,zhang2019man,xu2019multilevel,xiao2021natural,cao2021pursuit,yang2021deconfounded}: They match all moment proposals to the language query and select the one with the highest matching score as the prediction. 2) Anchor-free~\cite{lu2019debug,xiao2021boundary,chen2020rethinking,yang2020tree,yang2022video}: They directly predict the probability of being a boundary for each frame, or directly regress the temporal locations of the target moment. In this paper, we convert the grounding stage as a video grounding problem, and build on top of a SOTA model DEBUG~\cite{lu2019debug} by extending it into multiple segment outputs.

\begin{figure*}[t]
  \begin{center}
  \includegraphics[width=\linewidth]{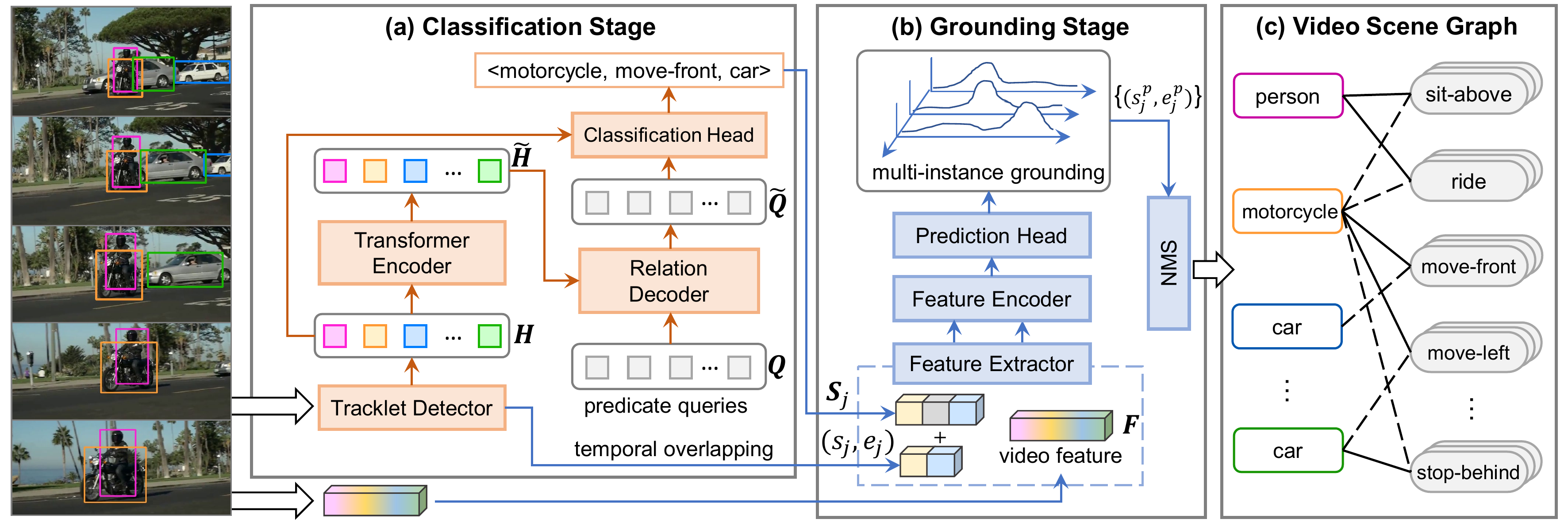}
  \end{center}
  \vspace{-1.5em}
  \caption{The overall pipeline of the proposed BIG model, which consists of a classification stage (a) and a grounding stage (b).}
  \label{fig:pipeline}
\end{figure*}

\section{Approach}

In this paper, we reformulate video scene graphs as temporal bipartite graphs. Given an entity category set $\mathcal{C}_e$ and predicate category set $\mathcal{C}_p$, a temporal bipartite graph is formally defined as $\mathcal{G} = (\mathcal{N}_e,\mathcal{N}_p,\mathcal{E})$, where $\mathcal{N}_e$, $\mathcal{N}_p$, and $\mathcal{E}$ denote the set of entity nodes, predicate nodes and edges, respectively. For each entity node $e_i \in \mathcal{N}_e$, it associates with an entity class $c^e_i \in \mathcal{C}_e$ and a time slot $(s^e_i, e^e_i)$. Similarly, for each predicate node $p_j \in \mathcal{N}_p$, it associates with a predicate class $c^p_j \in \mathcal{C}_p$ and a set of time slots $\{(s^p_{j,k}, e^p_{j,k})\}_{k=1}^{K_j}$. This multiple time slots setting implies that each predicate node has $K_j$ instances with the same category (happens $K_j$ times) in the same subject-object pair. $\mathcal{E} \subseteq \mathcal{N}_e \times \mathcal{N}_p \times \mathcal{C}_r$ is a set of mapping that maps an entity-predicate pair to a semantic role, 
\ie, $\mathcal{N}_e\times \mathcal{N}_p \rightarrow \mathcal{C}_r$, 
where $\mathcal{C}_r = \{$\texttt{subject}, \texttt{object}$\}$ is a semantic role set. The size of $\mathcal{N}_e$ and $\mathcal{N}_p$ are denoted as $n$ and $m$, respectively.

Under this new bipartite graph formulation, we propose a novel VidSGG model: \textbf{BIG}. The overview pipeline of BIG is illustrated in Figure~\ref{fig:pipeline}, which consists of two stages: classification stage (Sec.~\ref{Sec:classification_stage}) and grounding stage (Sec.~\ref{Sec:grounding_stage}).

\subsection{Classification Stage}\label{Sec:classification_stage}

\subsubsection{Overview}
The classification stage aims to classify the categories of all the nodes (\ie, entity and predicate), and the edges between them (\ie, the semantic roles). As shown in Figure~\ref{fig:pipeline}(a), the classification stage consists of four parts: a tracklet detector, an encoder, a decoder, and a classification head.

\textbf{Tracklet Detector.} Given a video, we use a pretrained tracklet detector to detect all tracklets in the video (denoted as entity set $\mathcal{N}_e$), and corresponding spatial-temporal locations, categories, and features. Specifically, for each entity $e_i \in \mathcal{N}_e$ with length $l_i$ (the number of frames), it is characterized by the bounding box coordinates $\bm{b}_i \in \mathbb{R}^{l_i\times 4}$, object category $c^e_i \in \mathcal{C}_e$, and a time slot $(s_i^e,e_i^e)$. We fix all the detection results (\ie, $\{\bm{b}_{i}\}$ and $\{c^e_i\}$) as the final predictions.

The tracklet feature $\bm{f}_i$ for each entity $e_i$ is a combination of appearance feature and spatial feature. The appearance feature $\bm{f}^a_i \in \mathbb{R}^{l_i \times d_a}$ is extracted at each frame based on the box locations by using RoIAlign~\cite{ren2015faster}. The spatial feature $\bm{f}^s_i \in \mathbb{R}^{l_i \times 8}$ is the concatenation of all box coordinates $\bm{b}_i$ and offsets $\Delta \bm{b}_i$, where $\Delta \bm{b}_{i,j}$ is the box coordinate offsets of two consecutive frames, \ie, $\Delta \bm{b}_{i,j} = \bm{b}_{i,j+1} - \bm{b}_{i,j}$. Then, the tracklet feature $\bm{f}_i \in \mathbb{R}^{l_i \times d_e}$ for entity $e_i$ is
\begin{equation}
 \bm{f}_i = \mathtt{Conv} \left[ \mathtt{MLP}_a(\bm{f}^a_i);\mathtt{MLP}_s(\bm{f}^s_i) \right],
\end{equation}
where $\mathtt{MLP}_a$ and $\mathtt{MLP}_s$ are two learnable MLPs, $[;]$ is a concatenate operation, and $\mathtt{Conv}$ is a 1D convolutional layer.

\textbf{Encoder.} Given entity features $\{\bm{f}_i\}$, the encoder aims to encode global context among all entities. Thus, we utilize the vanilla Transformer encoder~\cite{vaswani2017attention} as our encoder, where each layer consists of a multi-head self-attention ($\mathtt{MHSA}$) and a feed-forward network ($\mathtt{FFN}$). Since the sizes of the entity features are different, we first utilize a pooling operation to transform each feature $\bm{f}_i \in \mathbb{R}^{l_i \times d_e}$ to a fixed size feature $\bm{f}'_i \in \mathbb{R}^{l \times d_e}$, and use a MLP to mapping it into a vector $\bm{h}_{i} \in \mathbb{R}^{d_e}$. Then, we stack all entity features $\{\bm{h}_i\}$ into a matrix $\bm{H} \in \mathbb{R}^{n \times d_e}$, and feed the $\bm{H}$ into the encoder. The outputs of the encoder are contextualized features $\widetilde{\bm{H}} \in \mathbb{R}^{n \times d_e}$.

\textbf{Decoder.} The decoder is designed to predict the edges of the graph, and derive enhanced predicate representations for the following predicate classification. The inputs for the decoder is a fixed-size set of $m$ predicate queries with corresponding learnable embeddings $\bm{Q} \in \mathbb{R}^{m \times d_q}$. Each query is responsible for a predicate node in the bipartite graph. We built on top of the Transformer decoder and replace the original cross-attention with a \emph{Role-aware Cross-Attention} (RaCA). Therefore, each decoder layer is summarized as:
\begin{equation} \label{eq:decoder}
\begin{aligned}
   \bm{Q}'_{(i)} &= \mathtt{LNorm}(\bm{Q}_{(i)} + \mathtt{MHSA}(\bm{Q}_{(i)})), \\
   \bar{\bm{Q}}'_{(i)} & = \mathtt{RaCA}(\bm{Q}'_{(i)},\widetilde{\bm{H}},\widetilde{\bm{H}}), \\
   \bm{Q}''_{(i)} &= \mathtt{LNorm}(\bm{Q}'_{(i)} + \bar{\bm{Q}}'_{(i)}),  \\
   \bm{Q}_{(i+1)} &= \mathtt{LNorm}(\bm{Q}''_{(i)} + \mathtt{FFN}(\bm{Q}''_{(i)})),
\end{aligned}
\end{equation}
where $\mathtt{LNorm}$ is the layer normalization~\cite{ba2016layer}, $\bm{Q}_{(i)}$ is the input query embeddings of $i$-th decoder layer. The output of the last deocder layer is denoted as $\widetilde{\bm{Q}}$, \ie, the enchanced query embeddings. Meanwhile, the cross-attention matrix (inside RaCA module) of the last decoder layer is denoted as $\widetilde{\bm{A}}$, which can be regarded as a soft edge linkage of the bipartite graph. More details and discussions about the RaCA module (vs. original cross-attention) are in Sec.~\ref{Sec:RaCA}.

\begin{figure}[t]
   \begin{center}
   \includegraphics[width=\linewidth]{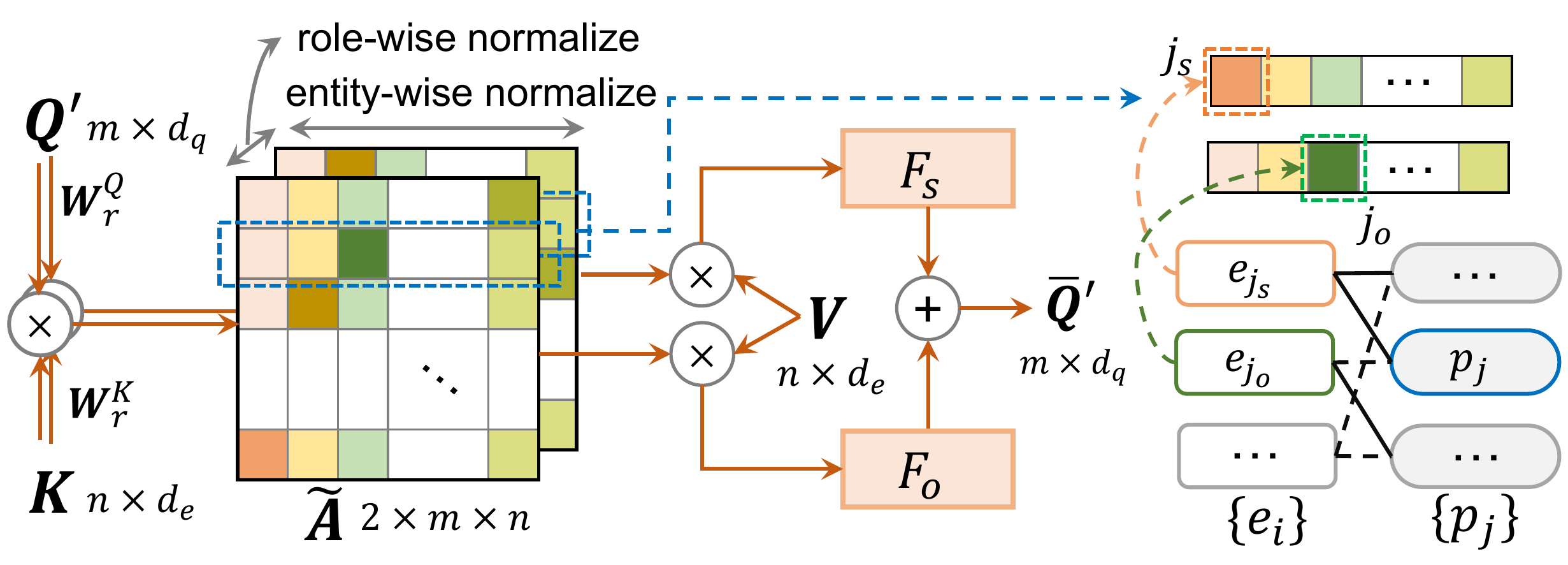}
   \end{center}
   \vspace{-2em}
   \caption{Illustration of the Role-aware Cross-Attention module.}
   \label{fig:RaCA}
\end{figure}

\textbf{Classification Head.} Given the query embeddings $\widetilde{\bm{Q}}$ and cross-attention matrix $\widetilde{\bm{A}}$, the classification head aims to classify the category of each query (\ie, predicate node). As shown in Figure~\ref{fig:RaCA},  $\widetilde{\bm{A}}$ has two channels which correspond to two different semantic roles in the bipartite graph. Based on $\widetilde{\bm{A}}$, we first derive the predicted subject and object for each predicate node $p_j$ by selecting the entity with the highest attention score in each channel, of which the indices are denoted as $j_s$ and $j_o$, respectively. Then, the classification feature $\bm{f}^p_j$ for predicate $p_j$ is a concatenation of three types of features: query embedding $\widetilde{\bm{q}}_j$, subject/object entity features $\bm{h}_{j_s}$ and $\bm{h}_{j_o}$, and word embeddings of subject/object entity categories, \ie, $\bm{f}^p_j = [\widetilde{\bm{q}}_j;\bm{h}_{j_s};\bm{h}_{j_o};\Pi(c^e_{j_s}); \Pi(c^e_{j_o})]$, where $\Pi(c^e_i) \in \mathbb{R}^{d_w}$ is the GloVe embedding~\cite{pennington2014glove} of object category $c^e_i$. Finally, the predicate category is classified by:
\begin{equation}\label{eq_cls}
    P(c_{j}^p) = \mathtt{Softmax}(\mathtt{MLP}_p(\bm{f}^p_j) + b_{c^e_{j_s},c^e_{j_o}}),
\end{equation}
where $\mathtt{MLP}_p$ is a MLP, and $b_{*,*}$ is the statistical prior of the relation triplet categories from the training set~\cite{zellers2018neural,teng2021target}.

\subsubsection{Role-aware Cross-Attention (RaCA)}\label{Sec:RaCA}
As shown in Figure~\ref{fig:RaCA}, the RaCA module aims to aggregate entity features from different semantic roles into query embeddings based on the cross-attention matrix. To distinguish different semantic roles (\ie, \texttt{subject} or \texttt{object}), we perform cross-attention for each semantic role separately, and then fuse these role-wise features with two non-linear transformations. Specifically, let $\bm{K}=\bm{V}=\widetilde{\bm{H}} \in \mathbb{R}^{n\times d_e}$ be the key and value matrix which are the output of the encoder, and $\bm{Q}' \in \mathbb{R}^{m \times d_q}$ be the query matrix which are the output from the first subnet in each decoder layer\footnote{For brevity, we omit subscripts $i$ in this subsection, \eg, $\bm{Q}'_{(i)} \to \bm{Q}'$.}  (cf. Eq.~\eqref{eq:decoder}). RaCA constructs a two-channel attention matrix $\bm{A}\in\mathbb{R}^{2 \times m\times n}$ and each channel of $\bm{A}$ is calculated as:
\begin{equation}
   \bm{A}_r = (\bm{Q}'\bm{W}_r^Q)(\bm{K}\bm{W}_r^K)^{\rm T} / \sqrt{d_e},
\end{equation}
where $\bm{W}_r^Q$, $\bm{W}_r^K$ are learnable weights and $r \in \{1,2\}$ represent the subject and object channel, respectively. In our formulation, since we assume each predicate query can only link to one entity in each role and each entity-predicate pair has one type of semantic roles at most. Thus, we normalize $\bm{A}$ along both the entity axis and the role axis, \ie,
\begin{align}
   \widetilde{\bm{A}}_{r,j,i} = \frac{\exp(\bm{A}_{r,j,i})}{\sum_{i'=1}^{n} \exp(\bm{A}_{r,j,i'})} 
   \times \frac{\exp(\bm{A}_{r,j,i})}{\sum_{r'=1}^{2} \exp(\bm{A}_{r',j,i})}. \label{doubel_softmax_A}
\end{align}
Then, we use two role-specific non-linear MLPs ($F_*$) to introduce role-wise 
distinctions into query embeddings,
\begin{align}\label{eq_Fr}
   \bar{\bm{Q}}' =  F_s(\widetilde{\bm{A}}_1\bm{V}) + F_o(\widetilde{\bm{A}}_2\bm{V}), ~ F_* : \mathbb{R}^{d_e} \mapsto \mathbb{R}^{d_q}, 
\end{align}
where $\bar{\bm{Q}}'$ is the output of the RaCA module (cf. Eq.~\eqref{eq:decoder}), which aggregates role-aware information from each entity.

\noindent\textbf{Discussions.} Compared to the plain cross-attention in original Transformer~\cite{vaswani2017attention}, RaCA explicitly learns the adjacency matrix of the bipartite graph based on role-wise normalization (cf. Eq.~(\ref{doubel_softmax_A})) and role-aware non-linear mappings ($F_*$ in Eq.~(\ref{eq_Fr})). Otherwise the adjacency matrix (or edge linkage) can not be modeled by the plain cross-attention module.

\subsection{Grounding Stage}\label{Sec:grounding_stage}
\begin{figure}[t]
   \begin{center}
   \includegraphics[width=\linewidth]{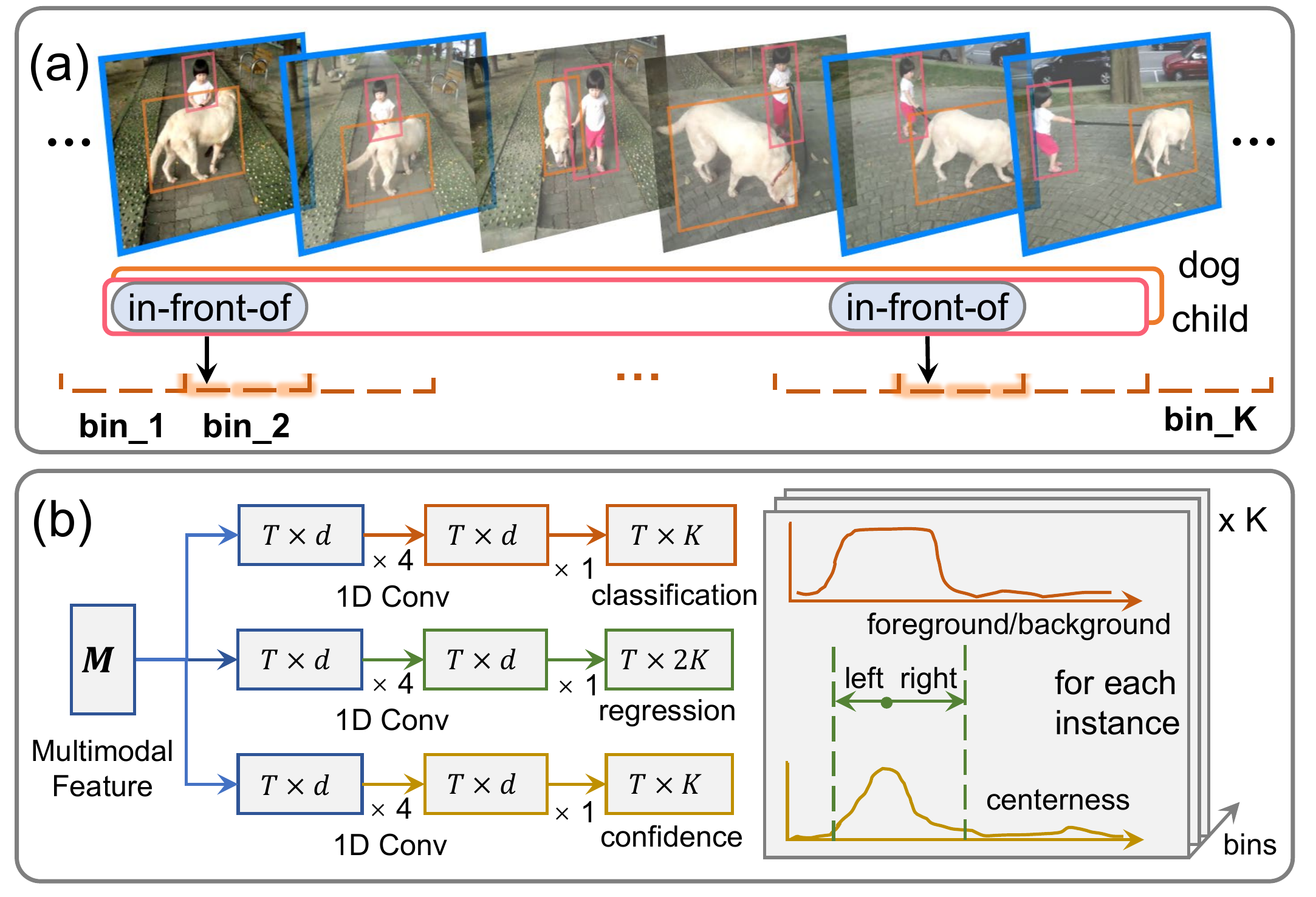}
   \end{center}
   \vspace{-1.5em}
   \caption{\textbf{(a)}: Illustration of label assignment in the multi-instance grounding. \textbf{(b)}: The overview of multi-instance grounding head.}
   \label{fig:grounding}
\end{figure}

The grounding stage aims to localize the temporal location of each predicted predicate node. So far, for each predicate node $p_j$, the classification stage have predicted its category $c^p_j$ and two linked entity tracklets: subject $e_{j_s}$ and object $e_{j_o}$. In this stage, we regard the predicate localization as a video grounding problem~\cite{gao2017tall}. Specifically, we treat the triplet categories sequence $(c^e_{j_s}, c^p_j, c^e_{j_o})$ (\eg, $\langle$\texttt{person}, \texttt{ride}, \texttt{motorcycle}$\rangle$ in Figure~\ref{fig:pipeline}(a)) as a language query, and extend an existing video grounding model DEBUG~\cite{lu2019debug} for multi-instance predicate localization. As shown in Figure~\ref{fig:pipeline}(b), this stage consists of three parts: a feature extractor, a feature encoder, and a multi-instance grounding head.

\textbf{Feature Extractor.} For the given video, we use a pretrained I3D~\cite{carreira2017quo_i3d} network to extract frame-level visual feature $\bm{F} \in \mathbb{R}^{T\times d_v}$, where $T$ is the number of whole video frames. For query $(c^e_{j_s}, c^p_j, c^e_{j_o})$ (refer to predicate node $p_j$), we initialize the query feature $\bm{S}_j = [\Pi(c^e_{j_s}), \Pi(c^p_j), \Pi(c^e_{j_o})]$, \ie, the GloVe embeddings of the triplet categories. Meanwhile, since each predicate only happen in the overlapping time of its subject and object, we use the temporal boundaries of this overlapping time as a prior feature to enhance $\bm{S}_j$, \ie, 
\begin{equation}
    \widetilde{\bm{S}}_j = \mathtt{MLP}_w(\bm{S}_j) + \mathtt{MLP}_t([s_j,e_j]), ~\widetilde{\bm{S}}_j \in \mathbb{R}^{3\times d_w},
\end{equation}
where $[s_j,e_j]\in \mathbb{R}^2$ are the overlapping boundaries of the subject and object linked to predicate node $p_j$. Note that only those predicate nodes referring to overlapped subject-object tracklets are used in the grounding stage. Visual features ${\bm{F}}$ is shared for all the queries with features $\{\widetilde{\bm{S}}_j\}_{j=1}^m$.

\textbf{Feature Encoder.} This encoder aims to model the interaction between the video feature $\bm{F}$ and all query features $\{\widetilde{\bm{S}}_j\}$. Specifically, we use the same feature encoder as DEBUG~\cite{lu2019debug}, which contains two parallel embedding encoders and a multi-modal attention layer. The output of the feature encoder is a fused multimodal feature $\bm{M} \in \mathbb{R}^{T\times d}$. We refer readers to the DEBUG~\cite{lu2019debug} paper for more details.

\textbf{Multi-instance Grounding Head.} Different from the existing video grounding task where each query only refers to a single segment, in VidSGG, a predicate category can happen multiple times between a same subject-object pair, \ie, each language query may refer to multiple segments (cf. Figure~\ref{fig:bipartite}). Since the number of time slots varies widely in different predicate nodes, it is difficult to directly predict a variable number of temporal segments for each query. Instead, we set $K$ bins for each language query. As shown in Figure~\ref{fig:grounding}(a), in the training stage, we divide the whole normalized video length evenly into $K$ intervals, referring to $K$ bins. Then, each bin is assigned with the target time slots centered in its interval\footnote{Although multiple targets may still fall into a same bin, such samples only account for a small proportion (details are in the appendix).}. In the test stage, all time slots predictions are processed by NMS to reduce false positives. Finally, the NMS operation results in $K_j$ time slots for the triplet query $(c_j^s,c_j^p,c_j^o)$, denoted as $\{(s_{j,k}^p,e_{j,k}^p)\}_{k=1}^{K_j}$.

Following DEBUG~\cite{lu2019debug}, we design three branches network for grounding: a classification subnet, a boundary regression subnet, and a confidence subnet (cf. Figure~\ref{fig:grounding}(b)). In particular, we extend the output channels of the last conv-layer to $K$ for the classification and confidence branch, and $2K$ for the regression branch (corresponding to $K$ bins).

\subsection{Training Objectives}\label{Sec:training}

\textbf{Classification Stage.} Since we fix all the tracklets from the detection backbone as the final entity nodes predictions, we only consider the training losses for classification of the edges and the predicate nodes. Let $\hat{\mathcal{N}}_p=\{\hat{p}_j\}_{j=1}^m$ be the predicted set of $m$ predicate nodes, and $\mathcal{N}_p^*$ be the ground-truth predicate set of size $m$ padded with $\varnothing$ (background). We adopt a one-to-one label assignment by finding a bipartite matching between $\hat{\mathcal{N}}_p$ and $\mathcal{N}_p^*$. Specifically, we search for a permutation of $m$ elements $\hat{\sigma}$ by optimizing the cost:
\begin{align} 
   \hat{\sigma} = \mathop{\arg\min}_{\sigma} \textstyle{\sum}_{j=1}^{m} \mathcal{L}_{\text{match}}( p^*_j,\hat{p}_{\sigma(j)}).
\end{align}
This matching problem can be computed efficiently with the Hungarian algorithm~\cite{munkres1957algorithms}, following prior work DETR~\cite{carion2020end}. 

The matching cost considers both predicate classification and edge prediction. Since all the entity nodes are fixed, the cost can be considered at the view of predicate nodes, and entity nodes are assigned to their ground-truths in advance (which is based on vIoU and the criterion is similar to that in Faster R-CNN~\cite{ren2015faster}). Thus, each predicate node can be described by its category and the two edges to subject/object. We denote $p^*_j = (c^{p*}_j,\bm{a}^*_j)$, where $c^{p*}_j$ is the predicate category (which may be $\varnothing$) and $\bm{a}^*_j \in \{0,1\}^{2\times n}$ is the $j$-th row of $\bm{A}^*$ (ground-truth adjacent matrix) for two channels. Note that $\bm{a}^*_{j,r,i}=0$ when the $i$-th entity has no ground-truth to match. For the predicted predicate with index $\sigma(j)$, the corresponding edges are described by $\hat{\bm{a}}_{\sigma(j)} \in \mathbb{R}^{2\times n}$, which is the $\sigma(j)$-th row of the predicted $\hat{\bm{A}}$ for two channels. With the above notations, the matching cost is defined as:
\begin{equation}
\begin{aligned}
   \mathcal{L}_{\text{match}}( p^*_j,\hat{p}_{\sigma(j)}) = &-\mathbf{1}_{\{c^{p*}_j\neq \varnothing\}} \log P(\hat{c}^p_{\sigma(j)} = c^{p*}_j) \\ 
   &+ \mathbf{1}_{\{c^{p*}_j\neq \varnothing\}} \lambda_\text{att} \mathcal{L}_\text{att}(\bm{a}^*_j,\hat{\bm{a}}_{\sigma(j)}),
\end{aligned}    
\end{equation}
where $\lambda_\text{att}$ is hyperparameter, and $\mathcal{L}_\text{att}$ is defined as a binary-cross entropy (BCE) loss, $\mathbf{1}_{\{\cdot\}}$ is an indicator function. After obtaining $\hat{\sigma}$, the loss $\mathcal{L}_c$ for classification stage consists of the matching loss between $(p^*_j,\hat{p}_{\hat{\sigma}(j)})$ pairs, and the background classification loss for other predicate nodes, \ie,
\begin{equation}
\small
\mathcal{L}_c =  
\textstyle{\sum}_{j} 
\mathcal{L}_{\text{match}}( p^*_j,\hat{p}_{\hat{\sigma}(j)}) -   
\textstyle{\sum}_{c^{p*}_j= \varnothing}
\log P(\hat{c}^p_{\hat{\sigma}(j)} = \varnothing).
\end{equation}

\textbf{Grounding Stage}. The grounding stage is trained separately from the classification stage, and we use ground-truth triplet categories for training. Following DEBUG~\cite{lu2019debug}, the training objectives consist of three losses for three respective branches. The total loss is averaged among all $K$ bins.

\section{Experiments}

\subsection{Datasets and Evaluation Metrics} \label{sec:datasets}

\textbf{Datasets.} We evaluated BIG on two benchmarks: 1) \textbf{VidVRD}~\cite{shang2017video}: It consists of 1,000 videos, which covers 35 object categories and 132 predicate categories. We used the official splits: 800 videos for training and 200 videos for test. 2) \textbf{VidOR}~\cite{shang2019annotating}: It consists of 10,000 videos, which covers 80 object categories and 50 predicate categories. We used official splits: 7,000 videos for training, 835 videos for validation, and 2,165 videos for test. Since the annotations of the test set are not released, we only evaluated the val set.

\textbf{Evaluation Metrics.} We evaluated BIG on two tasks: 1) \emph{Relation Detection} \textbf{(RelDet)}: It detects a set of visual relation triplets, and corresponding tracklets of subject and object. A detected triplet is considered to be correct if there is a same triplet tagged in ground-truth, and both subject and object tracklets have a sufficient vIoU (\eg, 0.5) with the ground-truth. We used mAP and Recall@K (R@K, K=50, 100) as metrics for RelDet. 2) \emph{Relation Tagging} \textbf{(RelTag)}: It only focuses on the precision of visual relation triplets and ignores the localization results of tracklets. For RelTag, we used Precision@K (P@K, K=1,5,10) as metrics.

\subsection{Implementation Details} \label{sec:implementation} 
\textbf{Tracklet Detector.} We utilized the video object detector MEGA~\cite{chen2020memory,gao2021video} with backbone ResNet-101~\cite{he2016deep} to obtain initial frame-level detection results, and adopted deepSORT~\cite{wojke2017simple} to generate object tracklets.

\textbf{Adapting BIG to VidVRD.} For each relation triplet in the training set of VidVRD~\cite{shang2017video}, we noticed that only a portion of ground-truth segments is annotated as foreground, which makes the annotated temporal boundaries unreliable for training. Therefore, we only used the classification stage of BIG for VidVRD, termed \textbf{BIG-C}. Consequently, the time slot for each predicate $p_j$ is calculated as the overlap of its subject and object, \ie, $(s^e_{j_s},e^e_{j_s}) \cap (s^e_{j_o},e^e_{j_o})$, and $K_j=1$.

More implementation details are left in the appendix.


\begin{figure*}[htbp]
    \begin{minipage}[c]{0.7\linewidth}
        \centering
        \captionsetup{type=table} 
        \addtolength{\tabcolsep}{-2.5pt}
        \begin{tabular}{l|cc|ccc|ccc}
        \hline
            \multirow{2}{*}{Models} & \multicolumn{2}{c|}{Features} & \multicolumn{3}{c|}{RelDet} & \multicolumn{3}{c}{RelTag} \\
            & \small{Visual} & \small{Motion} & \small{mAP} & \small{R@50} & \small{R@100} & \small{P@1} & \small{P@5} & \small{P@10} \\ 
        \hline
            \small{VidVRD}~\cite{shang2017video}$_{\emph{MM'17}}$ & \small{iDT} & \boldcheckmark & 8.58 & 5.54 & 6.37 & 43.00 & 28.90 & 20.80  \\
            \small{GSTEG}~\cite{tsai2019video}$_{\emph{CVPR'19}}$ & \small{iDT} & \boldcheckmark & 9.52 & 7.05 & 8.67 & 51.50 & 39.50 & 28.23 \\
            \small{VRD-GCN}~\cite{qian2019video}$_{\emph{MM'19}}$ & \small{iDT} & \boldcheckmark & 16.26 & 8.07 & 9.33 & 57.50 & 41.00 & 28.5  0 \\
            \small{MHA}~\cite{su2020video}$_{\emph{MM'20}}$ & \small{iDT} & \boldcheckmark & 19.03 & 9.53 & 10.38 & 57.50 & 41.40 & 29.45 \\
            \small{IVRD}~\cite{li2021interventional}$_{\emph{MM'21}}$ &   \small{RoI} & \boldcheckmark & 22.97 & 12.40 & 14.46 & 68.83 & 49.87 & 35.57 \\
            \small{VidVRD-II}~\cite{shang2021video}$_{\emph{MM'21}}$ & \small{RoI} & \boldcheckmark & 29.37 & 19.63 & 22.92 & 70.40 & 53.88 & 40.16 \\ 
            \small{Liu~\etal}~\cite{liu2020beyond}$_{\emph{CVPR'20}}$ & \small{RoI+I3D}$^\dagger$ & \boldcheckmark & 18.38 & 11.21 & 13.69 & 60.00 & 43.10 & 32.24 \\ 
            \small{Chen~\etal}\cite{Chen2021Social}$_{\emph{ICCV'21}}$ & \small{RoI+I3D} & \boldcheckmark & 20.08 & 13.73 & 16.88 & 62.50 & 49.20 & 38.45 \\
        \hline
            \small{Liu~\etal}~\cite{liu2020beyond}$_{\emph{CVPR'20}}$ & \small{RoI}$^\dagger$ & & 14.01 & 8.47 & \textbf{11.00} & 56.50 & 36.70 & 26.60 \\
            \small{TRACE}~\cite{teng2021target}$_{\emph{ICCV'21}}$  & \small{RoI} &  & 15.06 & 7.67 & 10.32 & --- & --- & --- \\
            \small{\textbf{BIG-C (Ours)}} & \small{RoI}$^\dagger$  &  & \cellcolor{mygray-bg}{\textbf{17.56}} & \cellcolor{mygray-bg}{\textbf{9.59}}  & \cellcolor{mygray-bg}{10.92} & \cellcolor{mygray-bg}{\textbf{56.50}} & \cellcolor{mygray-bg}{\textbf{44.30}} & \cellcolor{mygray-bg}{\textbf{32.35}} \\ 
        \hdashline
            \small{Liu~\etal}~\cite{liu2020beyond}$_{\emph{CVPR'20}}$  & \small{RoI+I3D}$^\dagger$ &  & 14.81 & 9.14 & \textbf{11.39} & 55.50 & 38.90 & 28.90 \\
            \small{TRACE}~\cite{teng2021target}$_{\emph{ICCV'21}}$ & \small{RoI+I3D} &  & 17.57 & 9.08 & 11.15 & \textbf{61.00} & \textbf{45.30} & \textbf{33.50} \\
            \small{\textbf{BIG-C (Ours)}} & \small{RoI+I3D}$^\dagger$ &  & \cellcolor{mygray-bg}{\textbf{17.67}} & \cellcolor{mygray-bg}{\textbf{9.63}} & \cellcolor{mygray-bg}{11.29} & \cellcolor{mygray-bg}{56.00} & \cellcolor{mygray-bg}{43.80} & \cellcolor{mygray-bg}{32.85} \\
        \hdashline
            \small{\textbf{BIG-C (Ours)}} & \small{RoI}$^\ddagger$ &  & \cellcolor{mygray-bg}{\textbf{26.08}} & \cellcolor{mygray-bg}{\textbf{14.10}} & \cellcolor{mygray-bg}{\textbf{16.25}} & \cellcolor{mygray-bg}{\textbf{73.00}} & \cellcolor{mygray-bg}{\textbf{55.10}} & \cellcolor{mygray-bg}{\textbf{40.00}} \\ 
        \hline
        \end{tabular}
        \vspace{-0.5em}
        \caption{Performance (\%) on VidVRD of SOTA methods.  \textbf{Visual}: $^\dagger$ means that these models use the same tracklets and features as Liu~\etal\cite{liu2020beyond}, and $^\ddagger$ means that these models use tracklets and features generated by MEGA. \textbf{Motion}: It refers to the relative motion feature of entity pairs~\cite{shang2021video}.}
        \label{tab:sota_vidvrd}
    \addtolength{\tabcolsep}{2.5pt}
    \end{minipage} \hfill
    \begin{minipage}[c]{0.29\linewidth}
        \includegraphics[width=\linewidth]{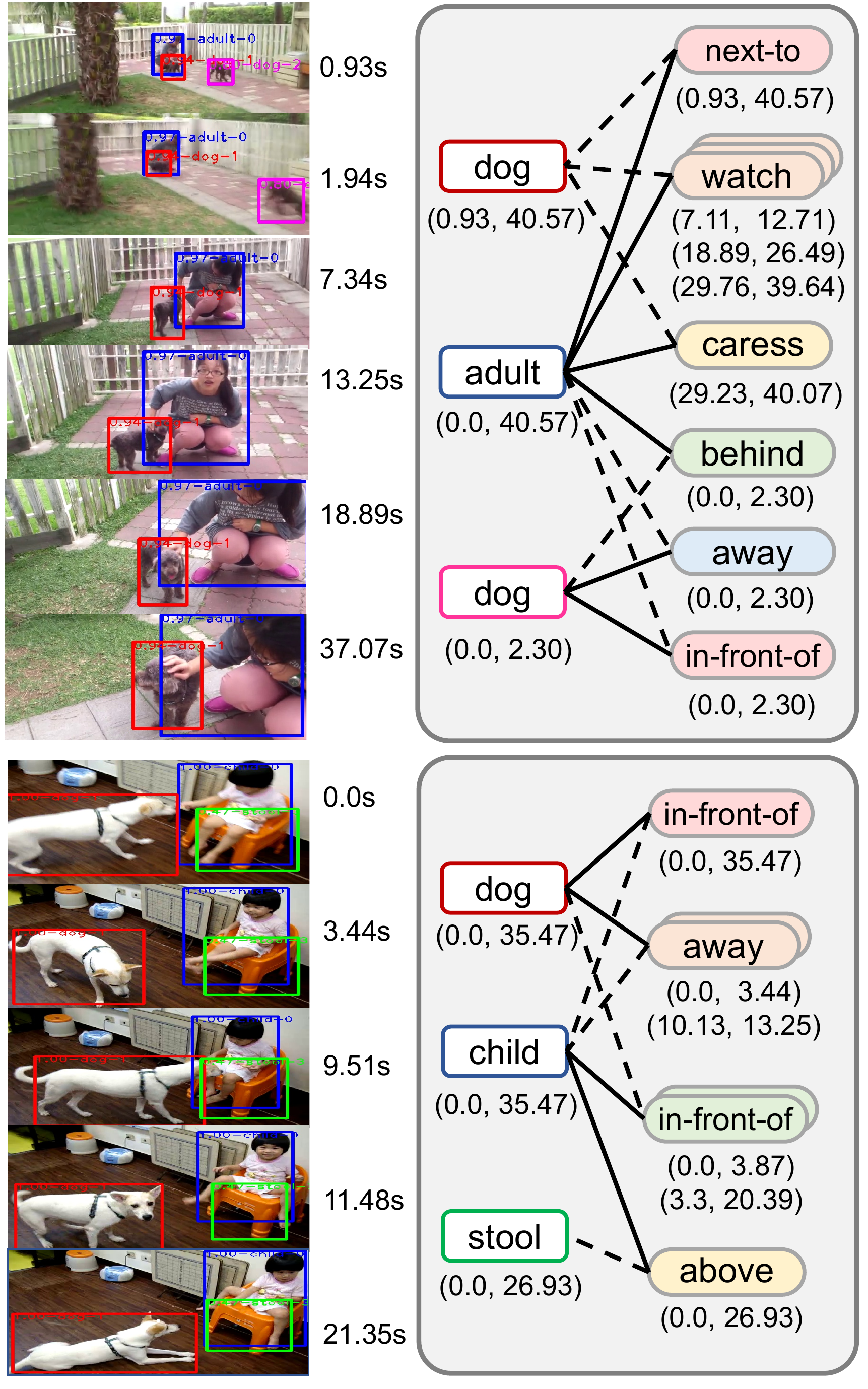}
        \vspace{-1.8em}
        \caption{Qualitative results on VidOR.}
        \label{fig:visualization}
    \end{minipage}
\end{figure*}

\subsection{Ablation Studies} \label{sec:ablation}

\textbf{Effectiveness of Classification-Then-Grounding.} We designed a baseline model to show the effectiveness of this framework and the two stages (classification \& grounding). Specifically, it directly classifies predicate categories of all tracklet pairs through multi-label classification, based on feature $\bm{f}'^p_j=[\bm{h}_{j_s};\bm{h}_{j_o};\Pi(c^e_{j_s}); \Pi(c^e_{j_o})]$ (cf. $\bm{f}^p_j$ in Eq.(\ref{eq_cls})), namely \textbf{Base-C}. Then, we apply the grounding stage to the Base-C, termed \textbf{Base}. All results are in Table~\ref{tab:cls_grd}. From this table, we can observe that even without the BIG model, the simple classification-then-grounding baseline (Base) still outperforms SOTA proposal-based model Sun~\etal~\cite{sun2019video}.

Furthermore, we reported the number of average relation candidates for the grounding stage (\#Cand.) in Table~\ref{tab:cls_grd} to demonstrate the effectiveness of each stage. For the classification stage, by comparing BIG-C with Base-C, 
we can observe that BIG-C outperforms Base-C on all metrics, especially with a large margin on RelTag while having fewer average relation candidates (135.4 vs. 482.1), which demonstrates the superiority of the encoder-decoder pipeline under the temporal bipartite graph formulation. For the grounding stage, we can observe that it can consistently improve detection mAP and recall for both two backbones (Base-C and BIG-C). The improvements of RelTag are slight because it doesn't consider the locations of relation triplets.

\addtolength{\tabcolsep}{-3.5pt} 
\begin{table}[t]
   \centering
   \begin{tabular}{l|ccc|cc|c}
    \hline
        \multirow{2}{*}{Models} & \multicolumn{3}{c|}{RelDet (\%)} & \multicolumn{2}{c|}{RelTag (\%)} & \multirow{2}{*}{\#Cand.} \\
        & \small{mAP} & \small{R@50} & \small{R@100} & \small{P@1} & \small{P@5} &  \\ 
    \hline
        Sun~\etal~\cite{sun2019video} & 6.56 & 6.89 & 8.83 & 51.20 & 40.73 & ---  \\ 
    \hline
        Base-C & 7.05 & 7.17 & 9.19  & 59.01 & 47.07 & 482.1 \\
        Base & 7.19 & 7.32 & 9.50  & 59.49 & 47.28 & 482.1 \\ 
        \textbf{BIG-C} & \cellcolor{mygray-bg}{8.29} & \cellcolor{mygray-bg}{7.92} & \cellcolor{mygray-bg}{9.65}  & \cellcolor{mygray-bg}{64.42} & \cellcolor{mygray-bg}{51.70} & \cellcolor{mygray-bg}{135.4} \\
        \textbf{BIG} & \cellcolor{mygray-bg}{\textbf{8.54}} & \cellcolor{mygray-bg}{\textbf{8.03}} & \cellcolor{mygray-bg}{\textbf{10.04}} & \cellcolor{mygray-bg}{\textbf{64.42}} & \cellcolor{mygray-bg}{\textbf{51.80}} & \cellcolor{mygray-bg}{\textbf{135.4}} \\ 
    \hline
   \end{tabular}
   \vspace{-0.5em}
   \caption{Ablations on effectiveness of different stages on VidOR.}
   \label{tab:cls_grd}
\end{table}
\addtolength{\tabcolsep}{3.5pt}

\textbf{Number of Predicate Queries.}
We compared BIG-C with different number of predicate queries ($m$) in Table~\ref{tab:num_queries}. It can be observed that more predicate queries always improve the final VidSGG performance, but also result in more computations (\eg, \#Cand.). To trade-off effectiveness and efficiency, we set $m$ = 192 for all the following experiments.

\textbf{Ablations on the RaCA Module.} We analyzed the impact of role-wise normalization (R-norm) and the two role-specific mappings ($F_*$) in the RaCA module on BIG-C. From the results in Table~\ref{tab:RaCA}, we can observe that both the R-norm and $F_*$ are important for the role-aware information encoding. Particularly, when both two techniques are used, the model achieves the best results, especially on P@1.

\addtolength{\tabcolsep}{-2.5pt} 
\begin{table*}[ht]
   \centering
   \begin{tabular}{l|c|cccc|ccc|ccc}
    \hline
        \multirow{2}{*}{Models} & \multirow{2}{*}{Detector} & \multicolumn{4}{c|}{Features} & \multicolumn{3}{c|}{RelDet} & \multicolumn{3}{c}{RelTag} \\
        &  &  \small{Visual} & \small{Lang} & \small{Motion} & \small{Mask} & \small{mAP} & \small{R@50} & \small{R@100} & \small{P@1} & \small{P@5} & \small{P@10}    \\ 
    \hline
         Liu~\etal~\cite{liu2020beyond}$_{\emph{CVPR'20}}$ & RefineDet & RoI+I3D$_r$  &          & \boldcheckmark & & 6.85 & 8.21 & 9.90 & 51.20 & 40.73 & ---  \\
        Chen~\etal\cite{Chen2021Social}$_{\emph{ICCV'21}}$ & Faster R-CNN & RoI+I3D$_r$  & \boldcheckmark & \boldcheckmark &  & 10.04 & 8.94 & 10.69 & 61.52 & 50.05 & 38.48 \\
        Chen~\etal\cite{Chen2021Social}$_{\emph{ICCV'21}}$ & Faster R-CNN & RoI+I3D$_r$  & \boldcheckmark & \boldcheckmark & \boldcheckmark & 11.21 & 9.99 & 11.94 & 68.86 & 55.16 & 43.40 \\
    \hline
        IVRD~\cite{li2021interventional}$_{\emph{MM'21}}$ & Faster R-CNN & RoI &  & \boldcheckmark & & 7.42 & 7.36 & 9.41 & 53.40 & 42.70 & ---  \\
        Chen~\etal\cite{Chen2021Social}$_{\emph{ICCV'21}}$ & Faster R-CNN & RoI &   & \boldcheckmark & & \textbf{8.93} & 7.38 & 9.22 & 56.89 & 44.76 & 34.07 \\
        VidVRD-II~\cite{shang2021video}$_{\emph{MM'21}}$ & Faster R-CNN & RoI &  & \boldcheckmark & & \textcolor{blue}{8.65} & \textbf{8.59} & \textbf{10.69} & 57.40 & 44.54 & 33.30 \\
        \textbf{BIG-C (Ours)} & MEGA & RoI &  &  & & \cellcolor{mygray-bg}{8.03} & \cellcolor{mygray-bg}{7.60} & \cellcolor{mygray-bg}{9.39} & \cellcolor{mygray-bg}{\textbf{62.25}} & \cellcolor{mygray-bg}{\textcolor{blue}{50.96}} & \cellcolor{mygray-bg}{\textcolor{blue}{40.30}}   \\ 
         \textbf{BIG (Ours)} & MEGA & RoI+I3D$_f$ &  &  & & \cellcolor{mygray-bg}{8.28} & \cellcolor{mygray-bg}{\textcolor{blue}{7.74}} & \cellcolor{mygray-bg}{\textcolor{blue}{9.82}} & \cellcolor{mygray-bg}{\textcolor{blue}{62.13}} & \cellcolor{mygray-bg}{\textbf{51.25}} & \cellcolor{mygray-bg}{\textbf{40.48}}   \\
    \hdashline
        VRU'19-top1\cite{sun2019video}$_{\emph{MM'19}}$ & FGFA & ---  & \boldcheckmark & \boldcheckmark &  & 6.56    & 6.89   & 8.83     & 51.20   & 40.73   & ---  \\
        MHA~\cite{su2020video}$_{\emph{MM'20}}$ & FGFA  & --- & \boldcheckmark & \boldcheckmark & &  6.59 & 6.35 & 8.05 & 50.72 & 41.56 & ---  \\
        VRU'20-top1\cite{xie2020video}$_{\emph{MM'20}}$ & CascadeRCNN & RoI & \boldcheckmark & \boldcheckmark & \boldcheckmark & \textbf{9.93} & \textbf{9.12} & --- & \textbf{67.43} & --- & ---  \\
        Chen~\etal\cite{Chen2021Social}$_{\emph{ICCV'21}}$ & Faster R-CNN  & RoI & \boldcheckmark & \boldcheckmark & &\textcolor{blue}{9.54} & \textcolor{blue}{8.49} & \textbf{10.17} & 59.24 & 47.24 & 35.99 \\
        \textbf{BIG-C (Ours)} & MEGA & RoI & \boldcheckmark  & & & \cellcolor{mygray-bg}{8.29} & \cellcolor{mygray-bg}{7.92}  & \cellcolor{mygray-bg}{9.65} & \cellcolor{mygray-bg}{64.42} & \cellcolor{mygray-bg}{\textcolor{blue}{51.70}} & \cellcolor{mygray-bg}{\textbf{41.05}}   \\ 
        \textbf{BIG (Ours)} & MEGA & RoI+I3D$_f$ & \boldcheckmark & &  & \cellcolor{mygray-bg}{8.54} & \cellcolor{mygray-bg}{8.03} & \cellcolor{mygray-bg}{\textcolor{blue}{10.04}} & \cellcolor{mygray-bg}{\textcolor{blue}{64.42}} & \cellcolor{mygray-bg}{\textbf{51.80}} & \cellcolor{mygray-bg}{\textcolor{blue}{40.96}} \\
    \hline
   \end{tabular}
   \vspace{-0.5em}
   \caption{Performance (\%) on VidOR of SOTA models. The \textbf{Best} and \textcolor{blue}{second best} are marked in according formats. \textbf{Visual}: I3D$_r$ and I3D$_f$ denote region-level and frame-level I3D features, respectively. \textbf{Lang}: The word embeddings of entity categories. \textbf{Motion}: It refers to the relative motion feature of entity pairs~\cite{shang2021video}. \textbf{Mask}: It means the localization mask of entities~\cite{xie2020video}. }
   \vspace{-0.5em}
   \label{tab:sota_vidor}
\end{table*}
\addtolength{\tabcolsep}{2.5pt}

\begin{table*}
       \parbox{.3\linewidth}{
       \centering
       \setlength{\tabcolsep}{2pt}
        \begin{tabular}{c|cc|cc|c}
           \hline
                \multirow{2}{*}{$m$} & \multicolumn{2}{c|}{\small{RelDet (\%)}} & \multicolumn{2}{c|}{\small{RelTag (\%)}} & \multirow{2}{*}{\#Cand.} \\
                & \small{mAP} & \small{R@50} & \small{P@1} & \small{P@5} &  \\
            \hline
                128 & 7.50 & 7.15 & 62.62 & 51.02 & 105.1  \\
                192 & 8.29 & 7.92 & 64.42 & 51.70 & 135.4  \\
                256 & 8.31 & 7.92 & 62.86 & 51.22 & 169.3  \\ 
            \hline
        \end{tabular}
        \vspace{-1.0em}
        \caption{Ablations of BIG-C for different number of predicate queries on VidOR.} 
        \label{tab:num_queries}
        }
        \hfill
       \parbox{.3\linewidth}{
       \centering
       \setlength{\tabcolsep}{2pt}
       \begin{tabular}{cc|cc|cc}
            \hline
                \multirow{2}{*}{R-norm} & \multirow{2}{*}{$F_*$} & \multicolumn{2}{c|}{\small{RelDet (\%)}} & \multicolumn{2}{c}{\small{RelTag (\%)}} \\
                & & \small{mAP} & \small{R@50} & \small{P@1} & \small{P@5} \\ 
            \hline
                & \boldcheckmark & 7.98 & 7.71 & 61.65 & 51.10  \\ 
                \boldcheckmark &      & 8.02 & 7.36 & 61.65 & 51.68  \\
                \boldcheckmark & \boldcheckmark & 8.29 & 7.92 & 64.42 & 51.70 \\ 
            \hline
       \end{tabular}
       \vspace{-1.0em}
       \caption{Ablations of BIG-C for the R-norm and $F_*$ of RaCA module on VidOR.}
       \label{tab:RaCA}
       }
       \hfill
       \parbox{.35\linewidth}{
       \centering
       \setlength{\tabcolsep}{2pt}
       \begin{tabular}{c|ccc|ccc}
            \hline
                \multirow{2}{*}{\#Bins} & \multicolumn{3}{c|}{\small{fR$_S$@K (\%)}} &   \multicolumn{3}{c}{\small{fR$_M$@K (\%)}} \\
                & \small{50} & \small{100} & \small{150} & \small{50} & \small{100} & \small{150} \\ 
            \hline
                1 & 12.96 & 15.59  & 16.76  & 5.53   & 6.86   & 7.46   \\
                5 & 13.07 & 15.83  & 17.26  & 5.75   & 7.20   & 8.05   \\
                10 & 13.04 & 15.89 & 17.61 & 5.75 & 7.30 & 8.25 \\ 
            \hline
       \end{tabular}
       \vspace{-1.0em}
       \caption{Ablations for multi-instance grounding with different number of bins on VidOR.}
        \label{tab:multi_ins_grd}
       }
\end{table*}

\textbf{Ablations on the Multi-instance Grounding.} We further investigated the influence of different number of bins in the multi-instance grounding. Since each predicate category of a same subject-object may have multiple instances, we regarded relation triplets with the same subject-object pair and predicate category as a sample. Each ground-truth sample can be partially hit with a \textbf{fraction recall} (\textbf{fR}), which is calculated as the fraction of hit relation triplets of each sample. For more precise, we evaluated fR@K for ground-truth samples with a single instance (fR$_S$) and multiple instances (fR$_M$), separately. From the results in Table~\ref{tab:multi_ins_grd}, we can observe that: 1) With the increase of $K$, corresponding fR@K increases on both single-instance and multi-instance samples. 2) Our multi-instance grounding (\eg, \#Bins=5,10) is more capable of improving the fR of predicates on multi-instance samples, \eg, the relative gains of fR$_M$ are larger than fR$_S$ (3.97\% (5.53$\rightarrow$5.75) vs. 0.61\% (12.96$\rightarrow$13.04)).

\subsection{Comparisons with State-of-the-Arts} \label{sec:sota}

\subsubsection{Performance on VidVRD}

\noindent\textbf{Settings.} For VidVRD, we compared our BIG-C with several state-of-the-art methods, which can be coarsely categorized into two groups: 1) Segment-proposal based methods: \textbf{VidVRD}~\cite{shang2017video}, \textbf{GSTEG}~\cite{tsai2019video}, \textbf{VRD-GCN}~\cite{qian2019video}, \textbf{MHA}~\cite{su2020video}, \textbf{IVRD}~\cite{li2021interventional}, \textbf{VidVRD-II}~\cite{shang2021video}, and \textbf{TRACE}~\cite{teng2021target}. 2) Tracklet-proposal based models methods: \textbf{Liu~\etal}~\cite{liu2020beyond} and \textbf{Chen~\etal}\cite{Chen2021Social}. For more fair comparisons, we also reported the results of BIG-C with the same features as~\cite{liu2020beyond}.

\noindent\textbf{Results.} All results are reported in Table~\ref{tab:sota_vidvrd}. From this table, we have following observations: 1) When using the MEGA backbone, BIG-C (with only RoI feature) beats most of the proposal-based methods even without the grounding stage. Particularly, we achieve a very high mAP (\ie, 26.08\%) and the highest P@1 (\ie, 73.00\%). 2) When using the same RoI feature as~\cite{liu2020beyond}, BIG-C outperforms TRACE~\cite{teng2021target} and Liu~\etal~\cite{liu2020beyond} on both RelDet and RelTag tasks, especially we achieve significant performance gains on mAP (17.56\% vs. 14.01\%). 3) When using the same RoI and I3D features as~\cite{liu2020beyond}, BIG-C achieves better results on mAP and R@50.

\subsubsection{Performance on VidOR} 

\noindent\textbf{Settings.} For VidOR, we compared our BIG (and BIG-C) with the state-of-the-art methods: \textbf{MHA}~\cite{su2020video}, \textbf{IVRD}~\cite{li2021interventional}, \textbf{VidVRD-II}~\cite{shang2021video}, \textbf{Liu~\etal}\cite{liu2020beyond}, \textbf{Chen~\etal}\cite{Chen2021Social}, and two top-1 methods~\cite{sun2019video,xie2020video} from Video Relation Understanding (VRU) Challenges. All results are reported in Table~\ref{tab:sota_vidor}. It is worth noting that we only use the frame-level I3D features (I3D$_f$) in BIG model (\ie, the grounding stage), while some works use more stronger region-level I3D features (I3D$_r$).

\noindent\textbf{Quantitative Results.} Due to the multifarious object detector backbones and features, it is difficult to fairly compare BIG (BIG-C) with these methods. From Table~\ref{tab:sota_vidor}, we can observe: 1) For BIG (BIG-C) without language feature, we achieve significant performance gains on RelTag (\eg, 51.25\% vs. 44.76\% on P@5), and also have competitive performance on RelDet. 2) For BIG (BIG-C) with language feature, we can also achieve comparable results on RelDet and RelTag, especially the highest P@5 (\ie, 51.80\%).


\noindent\textbf{Qualitative Results.} Figure~\ref{fig:visualization} shows some qualitative results that justifies the necessity of the grounding stage. Take $\langle$\texttt{dog}, \texttt{away}, \texttt{child}$\rangle$ for example: Without grounding, we can only use the temporal intersection of dog and child to approximate the time slot of away, \ie, (0, 35.47). Instead, with the help of multi-instance grounding, the time slots are predicted as (0, 3.44) \& (10.13, 13.25). Refer to the appendix for more details.


\section{Conclusions and Limitations}

In this paper, we pointed out three inherent drawbacks of the prevalent proposal-based framework, and proposed a new classification-then-grounding framework for VidSGG. Under this framework, we reformulated video scene graphs as temporal bipartite graphs, and proposed a novel VidSGG model BIG. We validated the effectiveness of BIG through extensive comparative and ablative experiments. 

\textbf{Limitations.} 1) Detecting long object tracklets in videos is still an open problem, and the fragmented tracklets may weaken the advantages of our framework, making it close to the proposal-based one.
2) Multi-instance grounding may not be suitable for some extreme situations where too many targets fall into the same bin (videos with dense relations).

\footnotesize \noindent\textbf{Acknowledgement.} This work was supported by National Key Research \& Development Project of China (2021ZD0110700), National Natural Science Foundation of China (U19B2043, 61976185), Zhejiang Natural Science Foundation (LR19F020002), Zhejiang Innovation Foundation (2019R52002), and Fundamental Research Funds for Central Universities.

\section*{Appendix}
\appendix

This supplementary document is organized as follows:
\begin{itemize}
    \item More explanations about the temporal bipartite graph are in Sec.~\ref{Sec:fig}.
    \item More detailed implementation details are in Sec.~\ref{supp:details}.
    \item Statistics about predicates with multiple instances for multi-instances grounding are in Sec.~\ref{supp:statistic}.
    \item More qualitative results are in Sec.~\ref{supp:visualize}.
    \item Potential negative societal impact are in Sec.~\ref{supp:social}.
\end{itemize}

\section{More Explanation about Temporal Bipartite Graph}\label{Sec:fig}

\begin{figure}[t]
    \begin{center}
    \includegraphics[width=\linewidth]{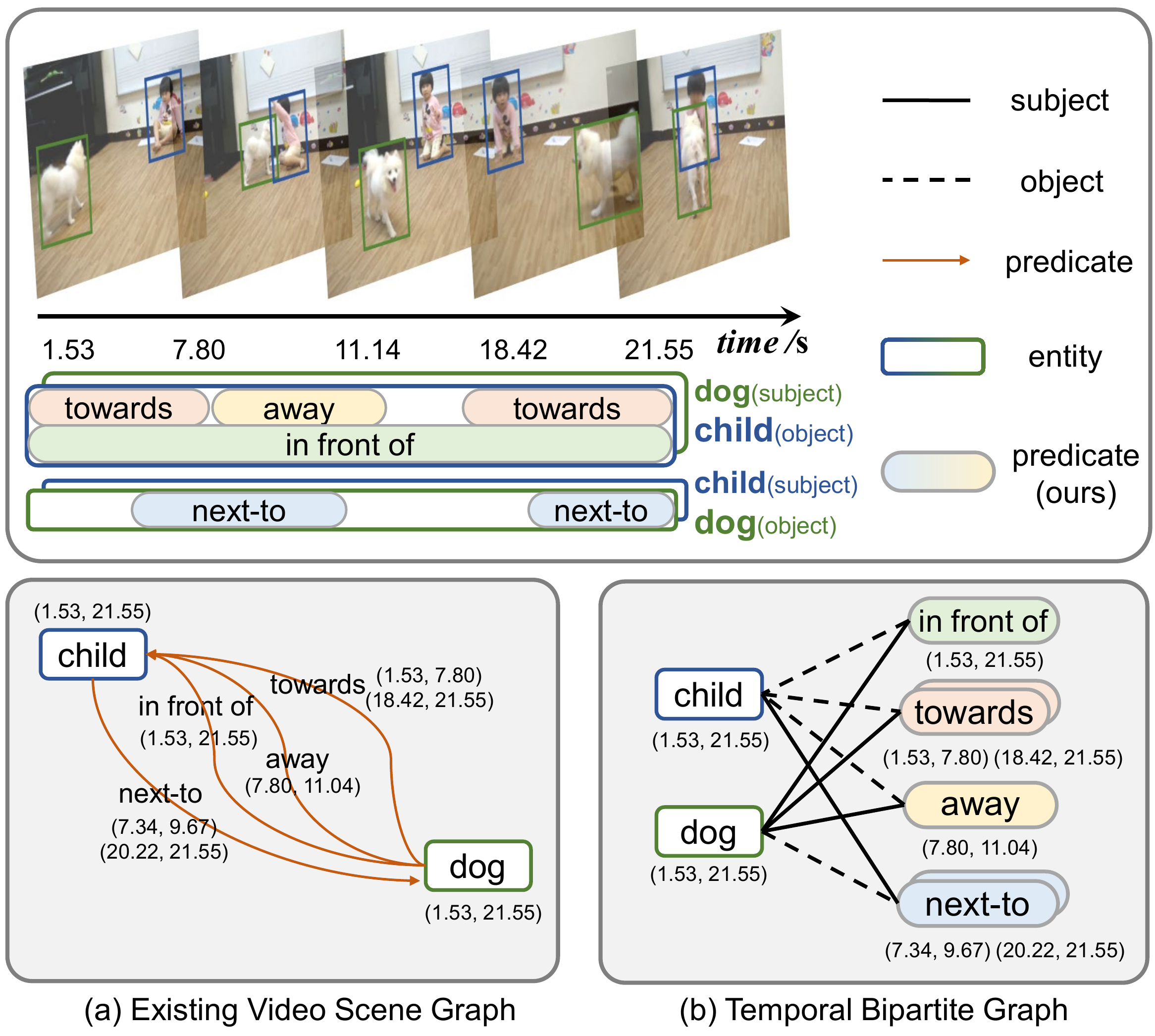}
    \end{center}
    \vspace{-1.5em}
    \caption{Existing video scene graph vs. temporal bipartite graph.}
    \label{fig_supp:bipartite}
\end{figure}

As shown in Figure~\ref{fig_supp:bipartite}, compared to existing video scene graphs (a), our temporal bipartite graph (b) have multiple advantages: 1) avoids exhaustively enumerating all entity pairs for predicate prediction; 2) is easier to model entity pairs with multiple predicates; and 3) can be easily extended to more general relations with more semantic roles (\eg, \texttt{instrument}~\cite{zareian2020weakly}).

\section{More Implementation Details} \label{supp:details}

\textbf{Tracklet Detector.} We utilized the video object detector MEGA~\cite{chen2020memory} with backbone ResNet-101~\cite{he2016deep} to obtain initial frame-level detection results, and adopted deepSORT~\cite{wojke2017simple} to generate object tracklets. This detector was trained on a video set and an image set. The video set is a set of downsampled videos from the training set, and the downsample rate were set to 5 and 32 for VidVRD and VidOR, respectively. The image set consists of images with the same categories as the VidSGG dataset, which are selected from the training and validation set of MSCOCO~\cite{lin2014microsoft} (for VidOR), or MSCOCO plus ILSVRC2016-DET~\cite{russakovsky2015imagenet} (for VidVRD).

\textbf{Parameter Settings.} 
The dimension of bounding box RoI feature $d_v$ was set to 1024, which was determined by MEGA. The dimensions of video I3D feature $d_I$ and word embedding $d_w$ were set to 1024 and 300, respectively. The hidden dimensions $d_q$ and $d_e$ were set to be 512. The output length $l_e$ of pooling operation was 4. The non-linear transforms $F_s,F_o$ are both MLPs. All the MLPs are two-layer FC networks with ReLU and the hidden dimension was 512. All bounding box coordinates and time slots are normalized to the range between (0, 1) w.r.t video size and video length, respectively. The loss factor was set as $\lambda = 30$. The parameters of DEBUG are set to the default settings~\cite{lu2019debug}.

\begin{figure*}[t]
    \begin{center}
    \includegraphics[width=\linewidth]{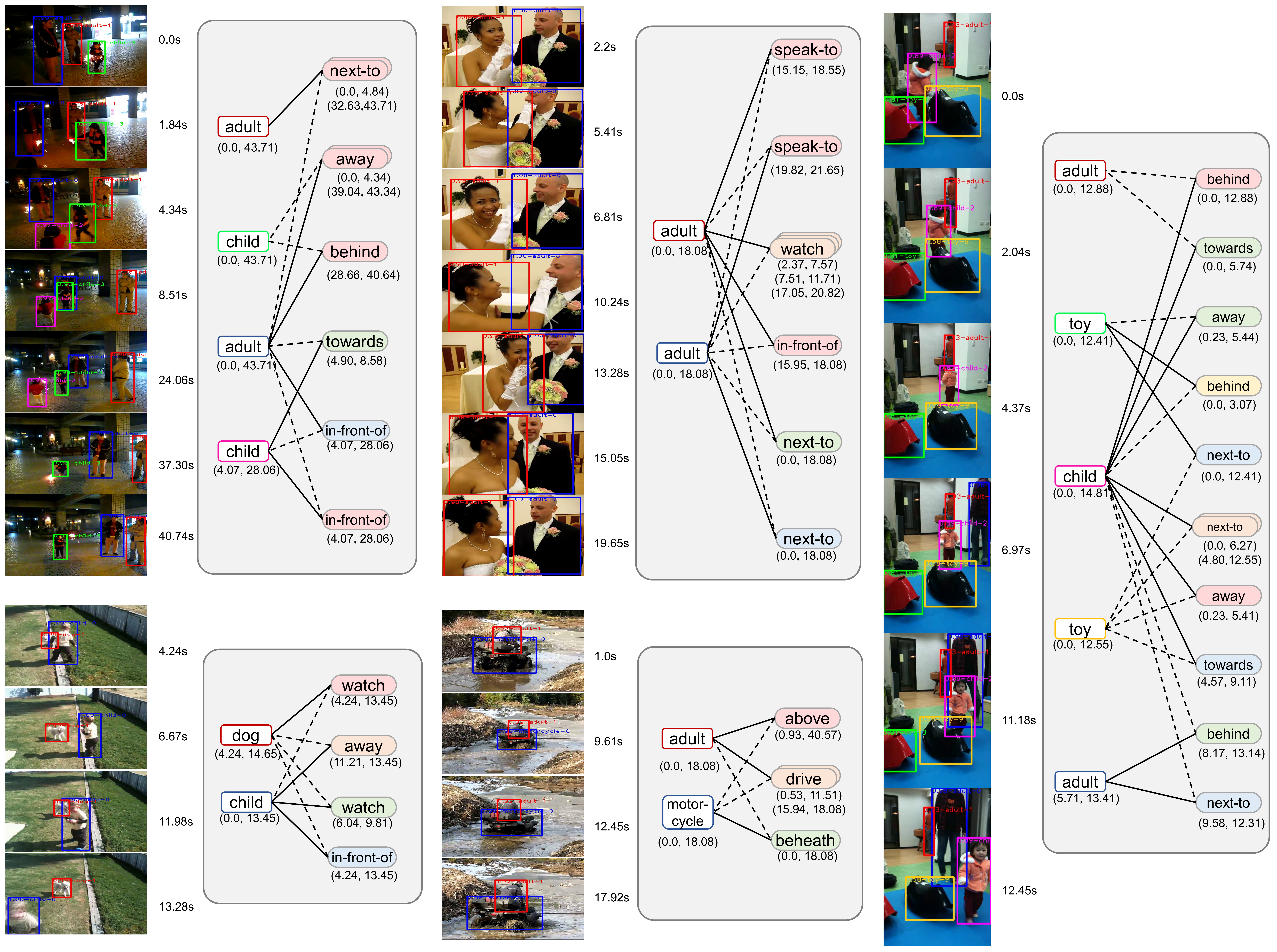}
    \end{center}
    \vspace{-1.5em}
    \caption{More qualitative results on VidOR validation set. The solid line and dash line represent the \texttt{subject} and \texttt{object} respectively.}
    \label{fig_supp:visualization}
\end{figure*}

\textbf{Training Details.} We trained our model for the classification stage and grounding stage separately.  For the classification stage, the model was trained Adam~\cite{kingma2014adam} for total 80/60 epochs with batch size 8/4 for VidVRD and VidOR, respectively. The learning rate was set to 1e-4 for VidVRD. For VidOR, the learning rate was set to 5e-5 in the first 50 epochs and 1e-5 in the last 10 epochs. For the grounding stage, the model was trained using ground-truth triplet categories in VidOR as language queries. It was trained by Adam~\cite{kingma2014adam} with 70 epochs and batch size of 8. The initial learning rate was set to 5e-5, and it decays 5 times in the 40-th and 60-th epoch.

\textbf{Inference Details.} In classification stage, following previous works~\cite{shang2017video}, we kept top-k triplet predictions (10 for VidVRD and 3 for VidOR) for each predicate node. Then, we filtered out duplicated triplets or triplets in which subject and object tracklets has no temporal overlap. In grounding stage, for each triplet query, we obtained $K$ time slots predictions through the multi-instance grounding, where each time slot prediction is associated with a score. Then, we filtered out these triplet queries whose highest score among all time slots is less than 0.2 (which might be false positives returned by the classification stage). For the remaining triplet queries,  we add the time slot of subject-object overlapping (with score 1.0) to get total $K$+1 instances for more robust prediction. Finally, we apply temporal NMS (with the threshold of 0.8) to these $K$+1 instance, which results in $K_j$ time slots for each predicate $p_j$. The tracklets for the relation triplet are cropped from  $e_{j_s},e_{j_o}$ according to the time slots of $p_j$, \ie, each $p_j$ is corresponding to $K_j$ relation triplets.



\section{Statistics for Multi-instance Predicates in VidOR}\label{supp:statistic}

\begin{figure}[t]
    \begin{center}
    \includegraphics[width=\linewidth]{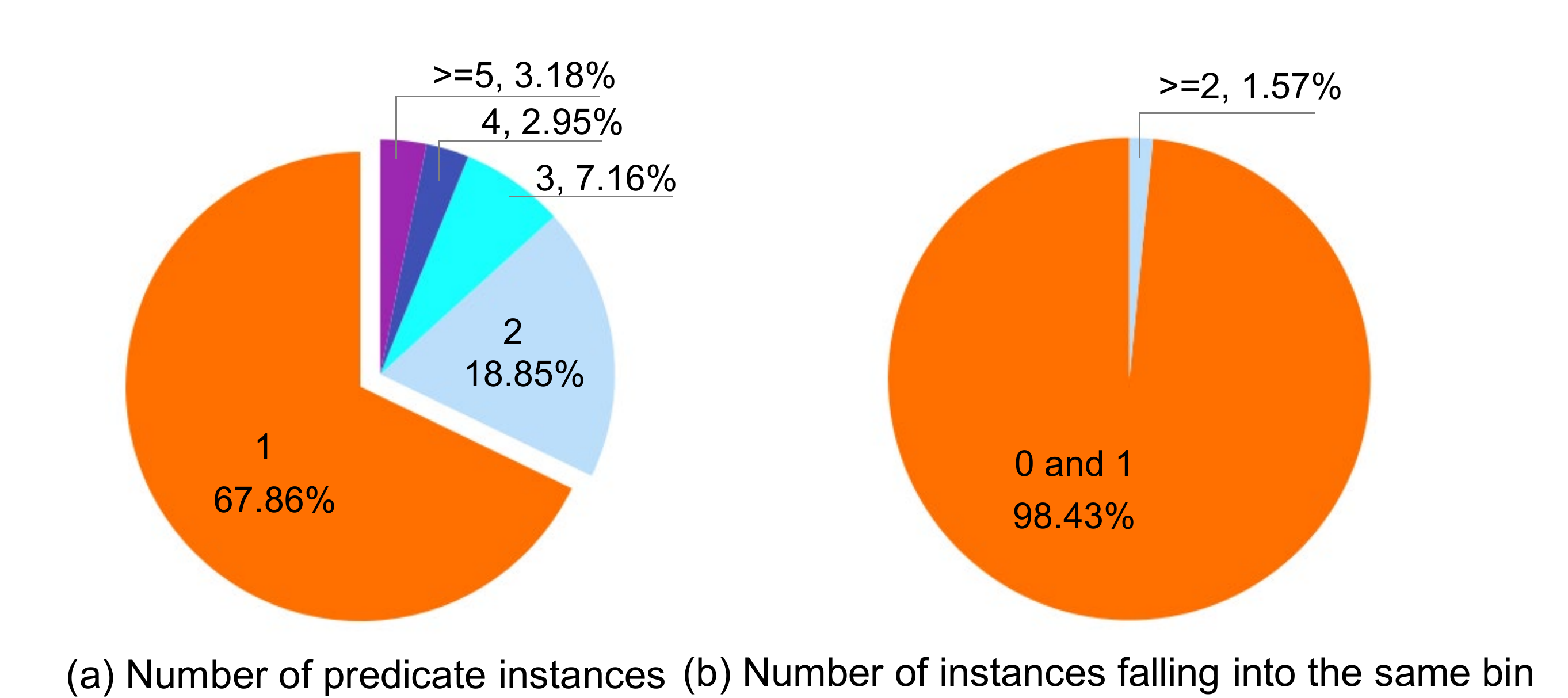}
    \end{center}
    \vspace{-1.5em}
    \caption{Statistics for multi-instance grounding from VidOR train set. \textbf{(a)}:Number of predicate instances with same category in a same subject-object pair. \textbf{(b)}:Number of instances falling into the same bin for predicates with multiple instances.}
    \label{fig_supp:pie_chart}
\end{figure}



We reported the distribution of predicates with single instance or multiple instances in Figure~\ref{fig_supp:pie_chart}(a). Here each sample is defined as the set of relation triplets with the same subject-object pair and predicate category. Then for those samples with multi-instance predicates, we reported the distribution of number of instances falling into the same bin for multi-instance grounding, as shown in Figure~\ref{fig_supp:pie_chart}(b), where each sample is a bin.


From Figure~\ref{fig_supp:pie_chart}(a), we can observe that although many predicates are single-instance, there are still around 32\% of predicates with multiple ground-truth instances. Thus,  the proposed multi-instance grounding is indispensable. Furthermore, Figure~\ref{fig_supp:pie_chart}(b) shows that only 1.57\% of bins are assigned with two instances or more, \ie, most of the bins are assigned with only one ground-truth target, which shows that our label assignment scheme is suitable for the multi-instance grounding.

\section{More Qualitative Results}\label{supp:visualize}

More qualitative results of our BIG model on VidOR are shown in Figure~\ref{fig_supp:visualization}.

\section{Potential Negative Societal Impact}\label{supp:social}

Our proposed \emph{classification-then-grounding} is a general framework of video scene graph generation (VidSGG), and there are no known extra potential negative social impact of our framework and BIG model. As for the challenging VidSGG task itself, there might be some wrong predictions (\eg, $\langle$\texttt{adult}, \texttt{kick}, \texttt{dog}$\rangle$). When VidSGG is applied to numerous down-stream tasks such as video captioning, video question answering, these wrong predictions might result in some ethical issues, \eg, a wrong caption says that a person is abusing a dog. To avoid the potential ethical issues, we can introduce some common sense knowledge into VidSGG models and design some rule-based methods to filter out those unreasonable relation triplets that involve ethical issues.

{\small
\bibliographystyle{ieee_fullname}
\bibliography{VidSGGbib}

\begin{thebibliography}{10}\itemsep=-1pt

\bibitem{anne2017localizing}
Lisa Anne~Hendricks, Oliver Wang, Eli Shechtman, Josef Sivic, Trevor Darrell,
  and Bryan Russell.
\newblock Localizing moments in video with natural language.
\newblock In {\em ICCV}, pages 5803--5812, 2017.

\bibitem{ba2016layer}
Jimmy~Lei Ba, Jamie~Ryan Kiros, and Geoffrey~E Hinton.
\newblock Layer normalization.
\newblock {\em arXiv}, 2016.

\bibitem{cao2021pursuit}
Meng Cao, Long Chen, Mike~Zheng Shou, Can Zhang, and Yuexian Zou.
\newblock On pursuit of designing multi-modal transformer for video grounding.
\newblock In {\em EMNLP}, 2021.

\bibitem{carion2020end}
Nicolas Carion, Francisco Massa, Gabriel Synnaeve, Nicolas Usunier, Alexander
  Kirillov, and Sergey Zagoruyko.
\newblock End-to-end object detection with transformers.
\newblock In {\em ECCV}, pages 213--229, 2020.

\bibitem{carreira2017quo_i3d}
Joao Carreira and Andrew Zisserman.
\newblock Quo vadis, action recognition? a new model and the kinetics dataset.
\newblock In {\em proceedings of the IEEE Conference on Computer Vision and
  Pattern Recognition}, pages 6299--6308, 2017.

\bibitem{chen2021human}
Long Chen, Zhihong Jiang, Jun Xiao, and Wei Liu.
\newblock Human-like controllable image captioning with verb-specific semantic
  roles.
\newblock In {\em CVPR}, pages 16846--16856, 2021.

\bibitem{chen2020rethinking}
Long Chen, Chujie Lu, Siliang Tang, Jun Xiao, Dong Zhang, Chilie Tan, and
  Xiaolin Li.
\newblock Rethinking the bottom-up framework for query-based video
  localization.
\newblock In {\em AAAI}, pages 10551--10558, 2020.

\bibitem{chen2021ref}
Long Chen, Wenbo Ma, Jun Xiao, Hanwang Zhang, and Shih-Fu Chang.
\newblock Ref-nms: breaking proposal bottlenecks in two-stage referring
  expression grounding.
\newblock In {\em AAAI}, pages 1036--1044, 2021.

\bibitem{chen2020counterfactual}
Long Chen, Xin Yan, Jun Xiao, Hanwang Zhang, Shiliang Pu, and Yueting Zhuang.
\newblock Counterfactual samples synthesizing for robust visual question
  answering.
\newblock In {\em CVPR}, pages 10800--10809, 2020.

\bibitem{chen2017sca}
Long Chen, Hanwang Zhang, Jun Xiao, Liqiang Nie, Jian Shao, Wei Liu, and
  Tat-Seng Chua.
\newblock Sca-cnn: Spatial and channel-wise attention in convolutional networks
  for image captioning.
\newblock In {\em CVPR}, pages 5659--5667, 2017.

\bibitem{chen2021counterfactual}
Long Chen, Yuhang Zheng, Yulei Niu, Hanwang Zhang, and Jun Xiao.
\newblock Counterfactual samples synthesizing and training for robust visual
  question answering.
\newblock {\em arXiv}, 2021.

\bibitem{chen2021reformulating}
Mingfei Chen, Yue Liao, Si Liu, Zhiyuan Chen, Fei Wang, and Chen Qian.
\newblock Reformulating hoi detection as adaptive set prediction.
\newblock In {\em CVPR}, 2021.

\bibitem{Chen2021Social}
Shuo Chen, Zenglin Shi, Pascal Mettes, and Cees G.~M. Snoek.
\newblock Social fabric: Tubelet compositions for video relation detection.
\newblock In {\em ICCV}, 2021.

\bibitem{chen2020memory}
Yihong Chen, Yue Cao, Han Hu, and Liwei Wang.
\newblock Memory enhanced global-local aggregation for video object detection.
\newblock In {\em CVPR}, pages 10337--10346, 2020.

\bibitem{cong2021spatial}
Yuren Cong, Wentong Liao, Hanno Ackermann, Bodo Rosenhahn, and Michael~Ying
  Yang.
\newblock Spatial-temporal transformer for dynamic scene graph generation.
\newblock In {\em ICCV}, pages 16372--16382, 2021.

\bibitem{gao2017tall}
Jiyang Gao, Chen Sun, Zhenheng Yang, and Ram Nevatia.
\newblock Tall: Temporal activity localization via language query.
\newblock In {\em ICCV}, pages 5267--5275, 2017.

\bibitem{gao2021video}
Kaifeng Gao, Long Chen, Yifeng Huang, and Jun Xiao.
\newblock Video relation detection via tracklet based visual transformer.
\newblock In {\em ACM MM}, pages 4833--4837, 2021.

\bibitem{he2016deep}
Kaiming He, Xiangyu Zhang, Shaoqing Ren, and Jian Sun.
\newblock Deep residual learning for image recognition.
\newblock In {\em CVPR}, pages 770--778, 2016.

\bibitem{hudson2019gqa}
Drew~A Hudson and Christopher~D Manning.
\newblock Gqa: A new dataset for real-world visual reasoning and compositional
  question answering.
\newblock In {\em CVPR}, pages 6700--6709, 2019.

\bibitem{johnson2015image}
Justin Johnson, Ranjay Krishna, Michael Stark, Li-Jia Li, David Shamma, Michael
  Bernstein, and Li Fei-Fei.
\newblock Image retrieval using scene graphs.
\newblock In {\em CVPR}, pages 3668--3678, 2015.

\bibitem{kingma2014adam}
Diederik~P Kingma and Jimmy Ba.
\newblock Adam: A method for stochastic optimization.
\newblock {\em arXiv}, 2014.

\bibitem{li2021interventional}
Yicong Li, Xun Yang, Xindi Shang, and Tat-Seng Chua.
\newblock Interventional video relation detection.
\newblock In {\em ACM MM}, pages 4091--4099, 2021.

\bibitem{lin2014microsoft}
Tsung-Yi Lin, Michael Maire, Serge Belongie, James Hays, Pietro Perona, Deva
  Ramanan, Piotr Doll{\'a}r, and C~Lawrence Zitnick.
\newblock Microsoft coco: Common objects in context.
\newblock In {\em ECCV}, pages 740--755, 2014.

\bibitem{liu2020beyond}
Chenchen Liu, Yang Jin, Kehan Xu, Guoqiang Gong, and Yadong Mu.
\newblock Beyond short-term snippet: Video relation detection with
  spatio-temporal global context.
\newblock In {\em CVPR}, pages 10840--10849, 2020.

\bibitem{liu2019joint}
Daqing Liu, Hanwang Zhang, Zheng-Jun Zha, Meng Wang, and Qianru Sun.
\newblock Joint visual grounding with language scene graphs.
\newblock {\em arXiv}, 2019.

\bibitem{lu2019debug}
Chujie Lu, Long Chen, Chilie Tan, Xiaolin Li, and Jun Xiao.
\newblock Debug: A dense bottom-up grounding approach for natural language
  video localization.
\newblock In {\em EMNLP}, pages 5144--5153, 2019.

\bibitem{munkres1957algorithms}
James Munkres.
\newblock Algorithms for the assignment and transportation problems.
\newblock {\em Journal of the society for industrial and applied mathematics},
  pages 32--38, 1957.

\bibitem{niu2021introspective}
Yulei Niu and Hanwang Zhang.
\newblock Introspective distillation for robust question answering.
\newblock In {\em NeurIPS}, 2021.

\bibitem{pennington2014glove}
Jeffrey Pennington, Richard Socher, and Christopher~D Manning.
\newblock Glove: Global vectors for word representation.
\newblock In {\em EMNLP}, pages 1532--1543, 2014.

\bibitem{qian2019video}
Xufeng Qian, Yueting Zhuang, Yimeng Li, Shaoning Xiao, Shiliang Pu, and Jun
  Xiao.
\newblock Video relation detection with spatio-temporal graph.
\newblock In {\em ACM MM}, pages 84--93, 2019.

\bibitem{ren2015faster}
Shaoqing Ren, Kaiming He, Ross~B Girshick, and Jian Sun.
\newblock Faster r-cnn: Towards real-time object detection with region proposal
  networks.
\newblock In {\em NeurIPS}, 2015.

\bibitem{russakovsky2015imagenet}
Olga Russakovsky, Jia Deng, Hao Su, Jonathan Krause, Sanjeev Satheesh, Sean Ma,
  Zhiheng Huang, Andrej Karpathy, Aditya Khosla, Michael Bernstein, et~al.
\newblock Imagenet large scale visual recognition challenge.
\newblock {\em IJCV}, pages 211--252, 2015.

\bibitem{shang2019annotating}
Xindi Shang, Donglin Di, Junbin Xiao, Yu Cao, Xun Yang, and Tat-Seng Chua.
\newblock Annotating objects and relations in user-generated videos.
\newblock In {\em ICMR}, pages 279--287, 2019.

\bibitem{shang2021video}
Xindi Shang, Yicong Li, Junbin Xiao, Wei Ji, and Tat-Seng Chua.
\newblock Video visual relation detection via iterative inference.
\newblock In {\em ACM MM}, pages 3654--3663, 2021.

\bibitem{shang2017video}
Xindi Shang, Tongwei Ren, Jingfan Guo, Hanwang Zhang, and Tat-Seng Chua.
\newblock Video visual relation detection.
\newblock In {\em ACM MM}, pages 1300--1308, 2017.

\bibitem{su2020video}
Zixuan Su, Xindi Shang, Jingjing Chen, Yu-Gang Jiang, Zhiyong Qiu, and Tat-Seng
  Chua.
\newblock Video relation detection via multiple hypothesis association.
\newblock In {\em ACM MM}, pages 3127--3135, 2020.

\bibitem{sun2019video}
Xu Sun, Tongwei Ren, Yuan Zi, and Gangshan Wu.
\newblock Video visual relation detection via multi-modal feature fusion.
\newblock In {\em ACM MM}, pages 2657--2661, 2019.

\bibitem{tamura2021qpic}
Masato Tamura, Hiroki Ohashi, and Tomoaki Yoshinaga.
\newblock Qpic: Query-based pairwise human-object interaction detection with
  image-wide contextual information.
\newblock In {\em CVPR}, 2021.

\bibitem{teng2021target}
Yao Teng, Limin Wang, Zhifeng Li, and Gangshan Wu.
\newblock Target adaptive context aggregation for video scene graph generation.
\newblock In {\em ICCV}, pages 13688--13697, 2021.

\bibitem{tsai2019video}
Yao-Hung~Hubert Tsai, Santosh Divvala, Louis-Philippe Morency, Ruslan
  Salakhutdinov, and Ali Farhadi.
\newblock Video relationship reasoning using gated spatio-temporal energy
  graph.
\newblock In {\em CVPR}, pages 10424--10433, 2019.

\bibitem{vaswani2017attention}
Ashish Vaswani, Noam Shazeer, Niki Parmar, Jakob Uszkoreit, Llion Jones,
  Aidan~N Gomez, Lukasz Kaiser, and Illia Polosukhin.
\newblock Attention is all you need.
\newblock In {\em NeuriIPS}, 2017.

\bibitem{wang2022crossformer}
Wenxiao Wang, Lu Yao, Long Chen, Binbin Lin, Deng Cai, Xiaofei He, and Wei Liu.
\newblock Crossformer: A versatile vision transformer hinging on cross-scale
  attention.
\newblock In {\em ICLR}, 2022.

\bibitem{wojke2017simple}
Nicolai Wojke, Alex Bewley, and Dietrich Paulus.
\newblock Simple online and realtime tracking with a deep association metric.
\newblock In {\em ICIP}, pages 3645--3649, 2017.

\bibitem{xiao2021natural}
Shaoning Xiao, Long Chen, Jian Shao, Yueting Zhuang, and Jun Xiao.
\newblock Natural language video localization with learnable moment proposals.
\newblock In {\em EMNLP}, 2021.

\bibitem{xiao2021boundary}
Shaoning Xiao, Long Chen, Songyang Zhang, Wei Ji, Jian Shao, Lu Ye, and Jun
  Xiao.
\newblock Boundary proposal network for two-stage natural language video
  localization.
\newblock In {\em AAAI}, pages 2986--2994, 2021.

\bibitem{xie2020video}
Wentao Xie, Guanghui Ren, and Si Liu.
\newblock Video relation detection with trajectory-aware multi-modal features.
\newblock In {\em ACM MM}, pages 4590--4594, 2020.

\bibitem{xu2019multilevel}
Huijuan Xu, Kun He, Bryan~A Plummer, Leonid Sigal, Stan Sclaroff, and Kate
  Saenko.
\newblock Multilevel language and vision integration for text-to-clip
  retrieval.
\newblock In {\em AAAI}, pages 9062--9069, 2019.

\bibitem{yang2020tree}
Xun Yang, Jianfeng Dong, Yixin Cao, Xun Wang, Meng Wang, and Tat-Seng Chua.
\newblock Tree-augmented cross-modal encoding for complex-query video
  retrieval.
\newblock In {\em ACM MM}, pages 1339--1348, 2020.

\bibitem{yang2021deconfounded}
Xun Yang, Fuli Feng, Wei Ji, Meng Wang, and Tat-Seng Chua.
\newblock Deconfounded video moment retrieval with causal intervention.
\newblock In {\em SIGIR}, pages 1--10, 2021.

\bibitem{yang2019auto}
Xu Yang, Kaihua Tang, Hanwang Zhang, and Jianfei Cai.
\newblock Auto-encoding scene graphs for image captioning.
\newblock In {\em CVPR}, pages 10685--10694, 2019.

\bibitem{yang2022video}
Xun Yang, Shanshan Wang, Jian Dong, Jianfeng Dong, Meng Wang, and Tat-Seng
  Chua.
\newblock Video moment retrieval with cross-modal neural architecture search.
\newblock {\em TIP}, 2022.

\bibitem{yuan2021closer}
Yitian Yuan, Xiaohan Lan, Xin Wang, Long Chen, Zhi Wang, and Wenwu Zhu.
\newblock A closer look at temporal sentence grounding in videos: Dataset and
  metric.
\newblock In {\em ACM MM workshop}, pages 13--21, 2021.

\bibitem{zareian2020weakly}
Alireza Zareian, Svebor Karaman, and Shih-Fu Chang.
\newblock Weakly supervised visual semantic parsing.
\newblock In {\em CVPR}, pages 3736--3745, 2020.

\bibitem{zellers2018neural}
Rowan Zellers, Mark Yatskar, Sam Thomson, and Yejin Choi.
\newblock Neural motifs: Scene graph parsing with global context.
\newblock In {\em CVPR}, pages 5831--5840, 2018.

\bibitem{zhang2019man}
Da Zhang, Xiyang Dai, Xin Wang, Yuan-Fang Wang, and Larry~S Davis.
\newblock Man: Moment alignment network for natural language moment retrieval
  via iterative graph adjustment.
\newblock In {\em CVPR}, pages 1247--1257, 2019.

\bibitem{zou2021end}
Cheng Zou, Bohan Wang, Yue Hu, Junqi Liu, Qian Wu, Yu Zhao, Boxun Li, Chenguang
  Zhang, Chi Zhang, Yichen Wei, et~al.
\newblock End-to-end human object interaction detection with hoi transformer.
\newblock In {\em CVPR}, 2021.

\end{thebibliography}
}

\end{document}